\documentclass[a4paper,fleqn]{cas-sc}

\usepackage[authoryear,longnamesfirst]{natbib}
\usepackage{booktabs}
\usepackage{listings}
\usepackage{xcolor}
\usepackage{url}

\lstdefinestyle{python}{
  language=Python,
  basicstyle=\ttfamily\small,
  keywordstyle=\color{blue},
  stringstyle=\color{red!60!black},
  commentstyle=\color{green!50!black},
  breaklines=true,
  frame=single,
  numbers=left,
  numberstyle=\tiny\color{gray},
}

\begin{document}
\let\WriteBookmarks\relax
\def\floatpagepagefraction{1}
\def\textpagefraction{.001}
\hfuzz=120pt

\shorttitle{AutoCause: Automating expert decisions in environmental time-series causal discovery}

\shortauthors{Marco Ruiz et al.}

\title[mode=title]{AutoCause: A Python framework that automates expert decisions in environmental time-series causal discovery}

\author[1]{Marco Ruiz}[orcid=0000-0002-5163-413X]
\cormark[1]
\ead{marco.rueda@tecnico.ulisboa.pt}
\credit{Conceptualization, Methodology, Software, Validation, Formal analysis, Writing -- original draft}

\affiliation[1]{organization={ISR-Lisbon, Instituto Superior T\'{e}cnico},
            addressline={Av. Rovisco Pais 1},
            city={Lisbon},
            postcode={1049-001},
            country={Portugal}}

\author[2]{Miguel Arana-Catania}[orcid=0000-0003-4277-0292]
\credit{Conceptualization, Supervision, Writing -- review and editing}

\affiliation[2]{organization={Digital Scholarship at Oxford, University of Oxford},
            city={Oxford},
            postcode={OX1 3BG},
            country={United Kingdom}}

\author[3]{David R. Ardila}[orcid=0000-0002-2564-8116]
\credit{Supervision, Writing -- review and editing}

\affiliation[3]{organization={Jet Propulsion Lab., Caltech},
            city={Pasadena},
            postcode={CA 91109},
            state={California},
            country={USA}}

\author[1]{Rodrigo Ventura}[orcid=0000-0002-5655-9562]
\credit{Supervision, Writing -- review and editing}

\cortext[1]{Corresponding author}

\begin{abstract}
Environmental time-series causal discovery requires expert decisions about method choice, conditional-independence tests, lag horizons, sample-size adequacy, multiple-testing control, and evidence interpretation. Applied inconsistently across datasets, these choices yield graphs that cannot be compared, reproduced, or audited. We present AutoCause, an open-source Python workflow that records each decision, derives defaults from an extended causal-audit module, and admits domain-informed overrides. The workflow wraps four established causal-discovery methods from three families, adds non-causal reference models, and grades links by method-count support. On 145 datasets from DGP-Atlas, TimeGraph, and a topology-derived CausalRivers reference, the methods recover complementary parts of the reference graphs. Majority-supported links are more precise than single-method links on the synthetic benchmarks but not against river topology. AutoCause converts inconsistent expert practice into an auditable, repeatable analysis; causal interpretation remains with the analyst. Available at \url{https://github.com/marcoruizrueda/autocause}.
\end{abstract}




\maketitle

\shortcites{delwiche2021fluxnetch4,runge2019inferring,arpit2023causalai,zheng2024causallearn,tetrad2018,smith2011netsim,krich2020estimating,pamfil2020dynotears,
moraffah2021causal,cheng2024cutsplus,zhang2021gcastle,zhang2025local,yang2026dycausal,
chen2023cuts,causaldynamics2025,LYU2026102435,marbach2010dream}

\section{Introduction}\label{sec:introduction}

Environmental models are commonly developed to answer questions that are causal in nature: which variables drive changes in an environmental response, which interactions are direct rather than mediated by other processes, which predictors are merely proxies for shared forcing, and which parts of a model are robust enough to support interpretation or management decisions.

In most of these systems, controlled experiments are impractical and the analyst must work with whatever multivariate record the monitoring infrastructure provides. A 10-year discharge series from a river network, a 3-year half-hourly flux record from a peatland, or a decadal satellite atmospheric column all share the same interpretive problem: the variables are autocorrelated, seasonally forced, nonlinearly coupled, and influenced by drivers that no sensor measured.
Under these conditions, simple association (Pearson or Spearman correlation) can be informative but insufficient: two variables may be highly correlated because both respond to the annual cycle, because one mediates the effect of another, or because a third unobserved process drives both.

Several families of causal-discovery methods have been developed to address this conflation, reviewed recently by \citet{brouillard2025grounding}, \citet{niu2024}, \citet{runge2023causal}, and \citet{runge2019inferring}. Each family rests on different mathematical assumptions about functional form, temporal memory, stationarity, and causal sufficiency \citep{assaad2022survey, moraffah2021causal, glymour2019review}. Constraint-based methods such as PCMCI+ require that a partial-correlation or mutual-information test can detect the conditional dependence, regression methods such as VARLiNGAM need non-Gaussian residuals for full identifiability, and information-theoretic approaches need large effective samples for their nearest-neighbour estimators. Violating any of these conditions produces a degraded or empty graph, and published comparisons find no single method that performs best across all data regimes \citep{assaad2022survey,nowack2020causal}. To support the method decision, \citet{runge2023causal} organize method choice through a questionnaire based on observable data characteristics, including linearity, stationarity, lag structure, and possible latent confounding. The questionnaire structures the decision, but the practitioner must still evaluate each characteristic and translate the result into a software configuration.

The questionnaire still leaves several steps to the analyst. A Tigramite workflow \citep{runge2019detecting}, for example, requires data assessment, conditional-independence (CI) test selection, a lag horizon, multiple-testing handling, and interpretation of method-specific output. Applying a fixed PCMCI+ and ParCorr configuration to seasonal or nonlinear benchmark data can produce low precision or recall \citep{ferdous2025timegraph}. \citet{brouillard2025grounding} identify the gap between methodological requirements and application settings as a barrier to using causal discovery on real data.

AutoCause was developed to close this gap by formalising the configuration and evaluation steps surrounding established causal-discovery algorithms. The framework invokes Tigramite for constraint-based methods and the LiNGAM library, but its contribution lies in the surrounding decision logic rather than in a new discovery algorithm. A single function call ingests a pandas DataFrame and records the six decisions defined above. Pre-discovery diagnostics from an extended causal-audit module \citep{ruiz2026causalaudit} inform method, CI-test, lag, and sample-size decisions together with any required preprocessing.

Once each method has completed its analysis and the corresponding $p$-values (where available) have undergone false-discovery-rate (FDR) correction, AutoCause counts how many of the four evaluated methods identify each candidate link and assigns the appropriate consensus-support tier. Dataset-level surrogate diagnostics are reported separately and are not incorporated into the tier assignment. The workflow generates one CSV file for each method, containing both the raw and FDR-corrected results, together with the consensus-tier edge list and cross-method comparison figures. When a reference graph is provided, the corresponding graph-recovery metrics are also reported. The execution configuration is stored alongside the generated outputs, allowing the analysis to be reproduced or repeated with modified settings. Apart from requiring a regularly sampled time axis, the interface does not assume a particular application domain. The same workflow can therefore be applied to river discharge measured every 6 hours, eddy-covariance fluxes sampled every 30 minutes, or satellite retrievals acquired at daily intervals.

The benchmarks used for the present evaluation comprise 145 datasets from three collections. DGP-Atlas \citep{ruiz2026causalaudit}, an atlas of data-generating processes (DGPs), contributes 97 synthetic series across 10 families of controlled assumption violations. TimeGraph \citep{ferdous2025timegraph} contributes 18 categories that probe linear, nonlinear, trend-seasonal, and missing-data regimes. CausalRivers \citep{stein2025causalrivers} contributes 30 five-station subgraphs from a real Bavarian discharge network evaluated against its physical topology. Other environmental domains, from atmospheric chemistry and marine biogeochemistry to ecological sensor arrays, remain outside the present scope and will need their own evaluation before transfer is claimed.

\paragraph*{Contributions.} This paper makes three contributions at different levels of the causal-discovery workflow.

At the software level, AutoCause provides the open-source pipeline that implements the workflow described above. Tigramite and the LiNGAM library supply the discovery algorithms; AutoCause adds the diagnostic, configuration, and evaluation layers that those libraries leave to the user. Every recorded default can be replaced with a domain-informed setting, and each method's raw output remains available for independent inspection. Among the releases compared in Table~\ref{tbl:frameworks}, none combines pre-discovery diagnostics, adaptive conditional-independence-test selection, method-count support grading, surrogate falsification, and non-causal reference models in one workflow. This integration is the software-level novelty of the paper.

At the pre-discovery level, the paper contributes three extensions to the original causal-audit module \citep{ruiz2026causalaudit}. VARLiNGAM enters the recommendation space for datasets whose risk scores indicate linear, non-Gaussian, and causally sufficient conditions. A diagnostic-extraction defect that silenced nonlinearity-aware and seasonality-aware routing has been corrected. The abstention rule has been revised so that seasonal data receive a deseasonalization recommendation rather than an unconditional rejection.

Finally, at the evaluation level, the benchmark experiments reported here cover graph recovery, method complementarity, consensus-support precision, and dataset-level surrogate behavior. DGP-Atlas is notable because, although its 97 synthetic graphs have been publicly available since 2025, no prior publication reports causal-discovery F1 scores on them. A single consensus-support analysis also spans all three collections, whose reference graphs come from a generative model, a structured benchmark, and a physical river topology. This common scale is what exposes the dependence of majority-support precision on the type of reference graph. The TimeGraph analysis compares the PCMCI+ results produced by AutoCause with those reported by \citet{ferdous2025timegraph}.

\section{Background and related work}\label{sec:background}

AutoCause wraps methods whose intellectual roots span vector-autoregressive (VAR) econometrics, information theory, and constraint-based graphical modeling. To situate these methods, we adopt the four-family taxonomy of \citet{assaad2022survey} and draw on the identifiability analysis of \citet{runge2023causal}, who map each family's assumptions to the environmental conditions under which it can or cannot recover causal structure.

\subsection{Why causal discovery matters for environmental modeling}\label{sec:why_causal_discovery}

Environmental monitoring produces multivariate time series from stream gauges, flux towers, atmospheric instruments, and ecological sensors \citep{runge2019inferring,delwiche2021fluxnetch4}. In these settings, researchers may ask whether drought response can be attributed to soil moisture, which measured variables precede methane flux changes in a peatland \citep{krich2020estimating}, or whether an upstream change is followed by a downstream water-quality response. Causal-discovery methods test which temporal dependencies remain after conditioning on other measured variables \citep{runge2023causal,runge2019inferring}.

Once estimated, such graphs support process identification by pairing candidate driver-response links with estimated lags. They can inform variable selection when direct-driver candidates are separated from seasonal or indirect associations, and they can motivate targeted field measurements or mechanistic-model refinement. Environmental applications range from climate-model evaluation and constrained projections \citep{nowack2020causal} to biosphere-atmosphere interactions \citep{krich2020estimating} and ecosystem methane dynamics \citep{delwiche2021fluxnetch4}.

These benefits do not imply that observational causal discovery can prove causality on its own. Its conclusions remain conditional on assumptions about sampling, temporal resolution, causal sufficiency, stationarity, and the adequacy of the chosen conditional independence or structural model. The practical value in environmental modeling therefore depends less on a single algorithm and more on a workflow whose assumptions are explicit, whose configuration is recorded, and whose outputs allow comparison across methodological paradigms. AutoCause fills this role by wrapping methods from the paradigms surveyed below within a reproducible pipeline that includes pre-discovery diagnostics, multi-method analysis, dataset-level surrogate diagnostics, and consensus-support tiering.

\subsection{Methods for time-series causal discovery}\label{sec:methods_taxonomy}

\citet{assaad2022survey} distinguish four families of time-series causal-discovery methods, a taxonomy adopted here. VAR-Granger \citep{granger1969investigating,toda1995statistical} and VARLiNGAM \citep{hyvarinen2010estimation} belong to the regression-based family. Transfer entropy and conditional mutual information form the information-theoretic family; nearest-neighbour conditional-mutual-information tests become more data-demanding as the conditioning dimension increases \citep{runge2018conditional}. Constraint-based methods such as PC, fast causal inference (FCI) \citep{spirtes2000causation}, PCMCI+ \citep{runge2020discovering}, and LPCMCI \citep{gerhardus2020high} prune links that lack conditional-independence support. Continuous-optimization methods such as DYNOTEARS encode graph constraints in an optimization objective \citep{pamfil2020dynotears}. Beyond these four families, learned-prior approaches provide a complementary option \citep{thumm2026causaltime_prior}. AutoCause includes methods from three of these families because their assumptions and finite-sample behavior differ.

The paradigms also differ in how they orient links, a distinction that governs the orientation comparison in Appendix~\ref{app:identifiability}. Regression-based and information-theoretic methods orient every link they report: VAR-Granger and transfer entropy by the time precedence of the lagged predictor, and VARLiNGAM additionally orients contemporaneous links through the non-Gaussianity of the residuals \citep{shimizu2006lingam}. Constraint-based methods orient lagged links by time order but identify contemporaneous links only up to the Markov equivalence class, returning them undirected when no unshielded collider or orientation rule fixes the direction \citep{runge2020discovering}. A symmetric association screen such as cross-correlation assigns no direction. These differences are intrinsic to the paradigms rather than to any implementation, so a comparison across methods reports adjacency first and treats orientation at the resolution each paradigm supports.

\subsection{Existing software frameworks}\label{sec:existing_software}

Given this diversity of methods and orientation conventions, a practitioner needs software that exposes not only the algorithms but also the configuration, preprocessing, and validation steps. Tigramite \citep{runge2019detecting} implements PCMCI+, LPCMCI, and several conditional-independence tests. The LiNGAM package implements discovery based on independent component analysis (ICA) for non-Gaussian models \citep{ikeuchi2023lingam}, and TETRAD \citep{tetrad2018} provides a Java graphical interface. Several frameworks expose multiple algorithms through a common application programming interface (API): Salesforce CausalAI \citep{arpit2023causalai} includes time-series and tabular methods, Causal-learn \citep{zheng2024causallearn} provides Python implementations of graphical-model algorithms, and gCastle \citep{zhang2021gcastle} emphasizes optimization-based structure learning.

Specialized extensions address narrower workflow elements. CausalFlow \citep{castri2023fpcmci} combines feature selection with PCMCI, VCDF \citep{vcdf2026} evaluates cross-temporal-fold consistency, and CausalNex is an open-source Python framework for Bayesian-network modeling that implements DYNOTEARS for structure learning \citep{causalnexsoftware,zheng2018dags, pamfil2020dynotears}. Bootstrap aggregation measures stability across repeated fits of one method \citep{debeire2024bagged}, and a related robust cross-validation procedure targets stability for validating agent-based-models \citep{rcv2026}. CausalMGM extends mixed graphical models to time-series constraint-based discovery \citep{andrews2019causalMGM}. Information-theoretic alternatives include the {tEDM} R package \citep{LYU2026102435} and the CausationEntropy software archive \citep{slote2025causationentropy}.

Tables~\ref{tbl:software_availability} and~\ref{tbl:frameworks} report the documented language, license, activity, and workflow capabilities of the compared releases. The feature comparison distinguishes multi-family execution, adaptive CI-test selection, method-count support grading, pre-discovery diagnostics, lag estimation, and non-causal reference models. Surrogate diagnostics are listed as a dataset-level falsification capability and are not treated as part of edge-tier assignment. AutoCause integrates all of these workflow steps around the discovery algorithms in a single reproducible call, which is why the upper block of Table~\ref{tbl:frameworks} shows checkmarks only in its column.

\begin{table}[pos=t]
\caption{Software availability of compared frameworks, based on cited papers and repository documentation checked on 11 July 2026.}\label{tbl:software_availability}
\small
\begin{tabular*}{\tblwidth}{@{}llll@{}}
\toprule
Framework & Language & License & Source \\
\midrule
AutoCause   & Python & AGPLv3+ & This work \\
Tigramite   & Python & GPL-3 & \citep{runge2019detecting} \\
CausalAI    & Python & BSD-3-Clause & \citep{arpit2023causalai} \\
CausalFlow  & Python & GPL-3 & \citep{castri2023fpcmci} \\
gCastle     & Python & Apache 2.0 & \citep{zhang2021gcastle} \\
causal-learn & Python & MIT & \citep{zheng2024causallearn} \\
VCDF        & Python & Apache 2.0 & \citep{vcdf2026} \\
CausalNex   & Python & Apache 2.0 & \citep{causalnexsoftware} \\
LiNGAM      & Python & MIT & \citep{ikeuchi2023lingam} \\
TETRAD      & Java & LGPL & \citep{tetrad2018} \\
\bottomrule
\end{tabular*}
\end{table}

\begin{table}[pos=t]
\caption{Capability comparison of time-series causal-discovery frameworks. $\checkmark$ means the feature is documented as built in, $\sim$ means partial or manual support, and -- means the feature was not found in the cited release. The lower block lists capabilities that AutoCause does not currently provide, included so that readers can judge what the framework omits.}
\label{tbl:frameworks}
\resizebox{\tblwidth}{!}{%
\begin{tabular}{@{}lccccccccc@{}}
\toprule
Capability & \rotatebox{60}{AutoCause} & \rotatebox{60}{Tigramite} & \rotatebox{60}{CausalAI} & \rotatebox{60}{CausalFlow} & \rotatebox{60}{gCastle} & \rotatebox{60}{VCDF} & \rotatebox{60}{CausalNex} & \rotatebox{60}{LiNGAM} & \rotatebox{60}{TETRAD} \\
\midrule
\multicolumn{10}{@{}l}{\textit{Capabilities integrated by AutoCause}} \\
Multi-paradigm         & $\checkmark$ & -- & $\sim$ & -- & $\sim$ & -- & -- & -- & $\sim$ \\
Adaptive CI-test       & $\checkmark$ & -- & -- & -- & -- & -- & -- & -- & -- \\
Evidence tiering       & $\checkmark$ & -- & -- & -- & -- & -- & -- & -- & -- \\
Falsification          & $\checkmark$ & $\sim$ & -- & -- & -- & $\sim$ & -- & -- & -- \\
Pre-discovery diag.    & $\checkmark$ & -- & -- & -- & -- & -- & -- & -- & -- \\
Correlation baseline   & $\checkmark$ & -- & -- & -- & -- & -- & -- & -- & -- \\
Predictive baseline    & $\checkmark$ & -- & -- & -- & -- & -- & -- & -- & -- \\
Auto $\tau_{\max}$     & $\checkmark$ & -- & -- & -- & -- & -- & -- & -- & -- \\
Paradigm diversity     & $\checkmark$ & -- & -- & -- & -- & -- & -- & -- & -- \\
TS-native              & $\checkmark$ & $\checkmark$ & $\checkmark$ & $\checkmark$ & $\sim$ & $\checkmark$ & $\sim$ & $\checkmark$ & $\sim$ \\
Latent confounders     & $\checkmark$ & $\checkmark$ & -- & -- & $\sim$ & -- & -- & -- & $\checkmark$ \\
\midrule
\multicolumn{10}{@{}l}{\textit{Where AutoCause is partial or absent}} \\
GPU acceleration       & --           & -- & $\checkmark$ & -- & $\checkmark$ & -- & $\checkmark$ & -- & -- \\
Neural causal methods  & --           & -- & $\checkmark$ & -- & $\checkmark$ & -- & -- & -- & -- \\
Continuous opt.\ (NOTEARS) & --       & -- & $\checkmark$ & -- & $\checkmark$ & -- & $\checkmark$ & -- & -- \\
Interventional data    & --           & -- & $\sim$       & -- & -- & -- & -- & -- & $\checkmark$ \\
Cyclic / feedback      & --           & -- & -- & -- & -- & -- & -- & -- & $\sim$ \\
Multi-realization (J-PCMCI+) & --     & $\checkmark$ & -- & -- & -- & -- & -- & -- & -- \\
Bootstrap aggregation  & --           & $\checkmark$ & -- & -- & -- & -- & -- & -- & -- \\
Graphical / no-code interface & --          & -- & $\checkmark$ & -- & -- & -- & -- & -- & $\checkmark$ \\
\bottomrule
\end{tabular}}
\end{table}

\subsection{Benchmarks and datasets}\label{sec:benchmarks_review}

\begin{table}[pos=t]
\caption{Benchmarks available for time-series causal discovery, grouped by how the reference graph is obtained. $\star$ marks the three used here. $^\ddagger$ indicates that the reference comes from a physical or engineered constraint external to the statistical model. Vars and $T$ report the number of variables and the series length; ODE and SDE denote ordinary and stochastic differential equations.}\label{tbl:datasets}
{\small
\begin{tabular*}{\tblwidth}{@{}p{2.5cm}p{2.2cm}p{1.15cm}p{1.3cm}p{2.4cm}p{4.6cm}@{}}
\toprule
Dataset & Domain & Vars & T & Challenge & Ref. \\
\midrule
\multicolumn{6}{@{}l}{\textit{Synthetic}} \\
CauseMe/C4C & Climate & 5--40 & 150--500 & Autocorrelation & \citep{runge2020causality4climate} \\
TimeGraph $\star$ & Generic & 4--20 & 1\,000 & Contemp. edges & \citep{ferdous2025timegraph} \\
CausalDynamics & ODE/SDE & Var. & Cont. & Stochastic & \citep{causaldynamics2025} \\
DGP-Atlas $\star$ & Synth. VAR & 5--8 & 500--1k & Violations & \citep{ruiz2026causalaudit} \\
\midrule
\multicolumn{6}{@{}l}{\textit{Semi-synthetic}} \\
CausalTime & Medicine, Air Quality,\newline traffic & 20--36 & 600--800 & Nonlinear & \citep{cheng2024causaltime} \\
DREAM3/4 & Gene net. & 10--100 & 21 & Short series & \citep{marbach2010dream} \\
NetSim & fMRI & 5--100 & 200 & HRF filter & \citep{smith2011netsim} \\
Krebs cycle & Biochem. & 16 & Var. & Reaction net & \citep{krebs_benchmark2025} \\
\midrule
\multicolumn{6}{@{}l}{\textit{Real-world}} \\
CausalRivers $\star$ $^\ddagger$ & Hydrology & 494 & 5 yr & Subgraphs & \citep{stein2025causalrivers} \\
CIPCaD (TEP)$^\ddagger$ & Industrial & 33 & 1.5k & Feedback & \citep{cipcad2022} \\
ESS Cryogenics$^\ddagger$ & Accelerator & 233 & Hours & High-dim & \citep{mogensen2024industrial} \\
Sachs (single-cell) & Biology & 11 & Var. & Mixed interv. & \citep{sachs2005causal} \\
\bottomrule
\end{tabular*}}
\end{table}

Available benchmarks span synthetic, semi-synthetic, and real-world domains (Table~\ref{tbl:datasets}). Synthetic collections provide a known generating graph. Semi-synthetic collections fit generators to observations and impose a known structure, so their interpretation remains conditional on the generator. Real-world benchmarks provide externally constructed reference structures, such as river topology for CausalRivers, a fault-propagation graph for CIPCaD, or a documented control architecture for ESS Cryogenics.

Each benchmark type has a different limitation. Synthetic data lack the measurement processes of field observations. Semi-synthetic data inherit assumptions from their generators. Real-world evaluations require part of the graph to be specified independently, and that reference may omit relevant processes. Many environmental applications lack such an external reference \citep{brouillard2025grounding}.

We evaluate on the three benchmarks marked $\star$ in Table~\ref{tbl:datasets}. They were chosen to cover three distinct evaluation regimes: a fully controlled synthetic setting where every link is known (DGP-Atlas), a structured synthetic benchmark with both lagged and contemporaneous ground truth that allows comparison against published outputs (TimeGraph), and a real environmental setting where the reference comes from river topology rather than from a generative model (CausalRivers). Below, each collection is described.

\textbf{DGP-Atlas} provides 500 synthetic VAR(1) processes organized in 10 families, each isolating a specific assumption violation (nonstationarity, structural breaks, irregular sampling, high persistence, latent confounders, seasonality, polynomial nonlinearity, non-Gaussian noise, mixed violations, and extreme cases). Ground truth is the VAR(1) coefficient matrix and all edges are lagged (lag 1), with no contemporaneous structure.

The \textbf{TimeGraph} dataset comprises 18 groups using four variables, having a lag of two time steps, and containing 1000 samples for each group. The dataset consists of four structural families. $A$ groups have a linear structure, while $B$ groups are linear structures but with nonlinearity of polynomials ($x^2$, $x^3$). $C$ groups have trends and seasons on top of the linear structure, and $D$ groups have missing data blocks. The confounded variant is marked by a \textit{C} suffix (A1C, B1C, C1C, D1C), in which one driving variable is removed from the observed set so that its influence appears as latent confounding.

Finally, the full \textbf{CausalRivers} benchmark provides two real-world river-network datasets, and we use RiversBavaria: 494 gauging stations recording water level at 15-minute intervals over 2019--2023, converted to discharge through rating curves maintained by the Bavarian Environment Agency. The reference graph is the directed river topology itself, where each edge represents a direct upstream-to-downstream station pair established from digital elevation models and field surveys. Because this reference encodes only channel connectivity, it does not capture shared rainfall, snowmelt routing, reservoir operations, tributary inflows, or other catchment processes that can generate statistical dependence between stations. A link that a discovery method reports but the topology omits is therefore scored as a false discovery, even though it may reflect a genuine hydrological coupling that the reference simply does not represent.

In order to investigate different types of methodological problems within this single river network, the benchmark uses three distinct sampling strategies for subgraphs. Sampling of randomly connected subgraphs of five stations is used to check how well the algorithms perform in complex topologies where no particular structure can be assumed. In root-cause chains, the longest directed path contains all five stations sequentially to check how well the algorithm distinguishes between the direct and indirect links by ignoring the latter through conditioning. For confounder subgraphs, in which there is at least one station with two or more downstream stations, the problem tested is how the methods respond to the presence of a common forcing variable. In all these cases, sampling is done without overlap between the rivers to ensure diversity in the sampling subgraphs.

\section{The AutoCause framework}\label{sec:framework}

The range of methods discussed above still requires practitioners to make six decisions before interpreting the results: which discovery method to run, which conditional-independence test to use, how to set the lag window, whether the available sample is adequate, how to control the false discovery rate, and how to grade the resulting evidence. AutoCause records a default or user-supplied value for each decision and logs the preprocessing actions that support it.

\subsection{Architecture and inputs/outputs}\label{sec:architecture}

\begin{figure}[pos=t]
  \centering
  \includegraphics[width=0.72\textwidth]{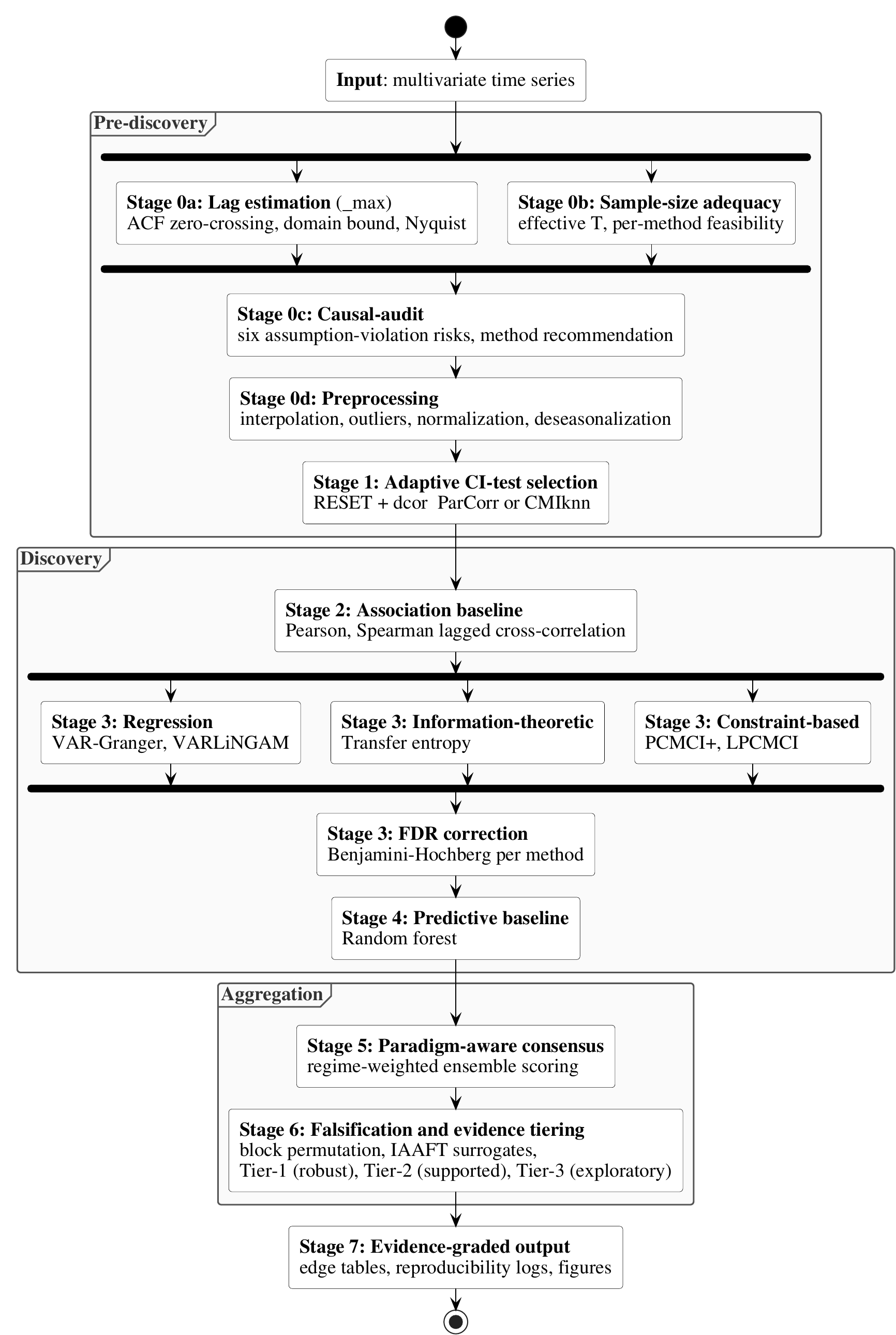}
  \caption{AutoCause pipeline. A complete call runs Stages 0--7. Optional stages can be selected subject to their input dependencies.}\label{fig:architecture}
\end{figure}

The AutoCause package is fully written in Python with a main experiment script around the single method \texttt{run\_causal\_discovery\_workflow}. There, the user provides a pandas DataFrame object with a time-based index and column names for all observable variables, an output directory, and, optionally, a dictionary specifying what stages/methods are to be executed. In the absence of the dictionary, all stages will operate with their default settings, while in the case of a partial dictionary, only the selected stages/methods will be executed. Within the function, there are seven stages of processing (Fig.~\ref{fig:architecture}) that use inputs from the preceding stages and generate outputs in corresponding directories.

The pre-discovery phase (Stage~0) estimates a search horizon and screens sample-size adequacy. The lag heuristic combines an autocorrelation zero-crossing, a user-supplied domain bound, and a data-availability constraint. The causal-audit module then maps diagnostic summaries for nonstationarity, irregular sampling, persistence, possible causal insufficiency, nonlinearity, and seasonality to risk scores in $[0,1]$. Its decision tree records a method recommendation and any indicated preprocessing.

Stage~1 selects the conditional-independence test. The Ramsey regression-equation specification error test (RESET) \citep{ramsey1969tests} and a distance-correlation comparison \citep{szekely2007measuring} assess nonlinearity, while a distributional diagnostic distinguishes standard and robust partial-correlation routes. The available routes are CMIknn, ParCorr, and RobustParCorr. Stage~2 provides a non-causal association reference by evaluating Pearson and Spearman correlations over lags $1,\ldots,\tau_{\max}$ and recording the minimum $p$-value and its lag.

Stage~3 runs four evaluated causal methods from three families: VAR-Granger and VARLiNGAM (regression), transfer entropy \citep{schreiber2000measuring,kraskov2004estimating} (information-theoretic), and PCMCI+ (constraint-based). LPCMCI is available but excluded from the benchmark loops under the stated wall-time budget. Where compatible $p$-values are available, the stage applies method-specific Benjamini--Hochberg correction and records the testing family. Stage~4 fits a random-forest predictive reference without assigning causal meaning to its importances.

Stage~5 records method agreement and assigns the method-count support tiers in Table~\ref{tbl:tier_definitions}. Stage~6 computes dataset-level iterative amplitude-adjusted Fourier transform (IAAFT) surrogate diagnostics \citep{schreiber2000iaaft} and temporal-stability summaries. These diagnostics do not define the edge tiers. Stage~7 writes configurations, method outputs, metrics, and figures. A supplied reference graph enables graph-recovery scoring. Otherwise, the workflow retains the discovered links and diagnostic evidence.

Optional stages can be selected subject to their input dependencies. Each dataset directory separates raw and corrected method outputs, consensus-support labels, diagnostic figures, and the executed configuration. Table~\ref{tbl:outputs} maps the principal artifacts to their interpretive purpose. A complete file manifest is included in the versioned paper release, and Section~\ref{sec:benchmarks} identifies the files used for the reported results.

\subsection{How to use it in practice}\label{sec:howto}

A practitioner does not need to invoke every pipeline stage at once. The interpretation sequence described here separates data assessment from causal-discovery output. The \texttt{audit-only} option returns the assumption-risk assessment and method recommendation before discovery begins. The recorded diagnostics cover stationarity, sample-size adequacy, functional form, seasonality, possible causal insufficiency, and the temporal dependence horizon. Association and predictive references can then show what structure is present before causal methods test which links remain after conditioning and method-specific multiple-testing control.

The lagged-correlation reference in Stage~2 provides the link-level results used in the association-to-causation comparison in Appendix~\ref{app:selectivity_progression}. That analysis examines how much of the detected structure remains as stronger conditioning requirements are introduced. Because lagged correlation tests marginal association, it may recover direct links, indirect paths, and dependencies induced by shared drivers.

The random-forest reference in Stage~4 fits one \texttt{RandomForestRegressor} for each target variable using lagged values of the remaining variables as predictors. Permutation importance identifies predictors whose shuffled lagged values reduce out-of-sample predictive performance. This model serves as a flexible predictive reference but it is neither a causal-discovery method nor a formal upper bound on predictability. Agreement with a causal method indicates that the corresponding lagged predictor contains predictive information, nevertheless, it does not establish causal direction. Both non-causal references are excluded from consensus tiering.

\begin{table*}[pos=t]
\caption{Artifacts written per dataset by \texttt{run\_causal\_discovery\_workflow} and the practitioner question each answers, grouped by pipeline phase. \textbf{Non-causal baselines} comprise lagged correlation (Stage~2) and the predictive baseline (Stage~4). In \textbf{Causal discovery}, \texttt{<m>} spans all causal methods (Granger, transfer entropy, PCMCI+, LPCMCI, VARLiNGAM).}\label{tbl:outputs}
\small\renewcommand{\arraystretch}{0.80}%
\begin{tabular*}{\tblwidth}{@{}>{\raggedright\arraybackslash}p{8.05cm}p{0.85cm}>{\raggedright\arraybackslash}p{7.6cm}@{}}
\toprule
Artifact & Stage & \textbf{Practitioner question answered} \\
\midrule
\multicolumn{3}{@{}l}{\textbf{Pre-discovery diagnostics}} \\
\url{causal_audit/figures/prediscovery_summary} & 0c & Data risks, method and CI-test recommendation \\
\url{causal_audit/figures/stationarity_diagnostic} & 0c & Stationarity over the record \\
\url{causal_audit/figures/sample_size_adequacy} & 0b & $T_\text{eff}$ adequacy for the selected method \\
\url{causal_audit/figures/spectral_density} & 0c & Seasonality detection and deseasonalization flag \\
\url{causal_audit/figures/correlation_and_lagged} & 0c & Pairwise association at each lag \\
\url{causal_audit/figures/assumption_deep_dive} & 0c & Violation magnitude per assumption \\
\url{causal_audit/figures/dependency_network} & 0c & Marginal dependency structure \\
\url{causal_audit/audit_evidence.json} & 0c & Raw diagnostic test values (machine-readable) \\
\url{causal_audit/diagnostic_tables/*.csv} & 0c & Stationarity, persistence, irregularity, and confounding results per pair \\
\url{causal_audit/risk_profile.json} & 0c & Six calibrated risk scores (machine-readable) \\
\url{causal_audit/recommendation_policy.json} & 0c & Recommended method, CI test, and preprocessing \\
\url{sample_size_adequacy.json} & 0b & $T_\text{eff}$ per method, viable methods \\
\url{power_analysis.json} & 0b & Minimum detectable effect size \\
\midrule
\multicolumn{3}{@{}l}{\textbf{Lag estimation}} \\
\url{tau_max_estimation.json} & 0a & Estimated $\tau_{\max}$ and bounding constraint \\
\midrule
\multicolumn{3}{@{}l}{\textbf{Non-causal baselines}} \\
\url{method/correlation/1-raw/results_correlation.csv} & 2 & Lagged Pearson and Spearman with best-lag selection \\
\url{figures/correlation/correlation_heatmap.png} & 2 & Comparison of Pearson, Spearman, Kendall, and distance correlation \\
\url{figures/correlation/method_comparison.png} & 2 & Agreement and disagreement across four correlation measures \\
\url{figures/correlation/partial_correlation_network.png} & 2 & Associations surviving control for other variables \\
\url{method/predictive_baseline/1-raw/results_predictive_baseline.csv} & 4 & Predictive importances with permutation p-values \\
\url{figures/per_method/predictive_baseline_graph.svg} & 4 & Predictive importance network \\
\url{figures/per_method/predictive_baseline_lags.svg} & 4 & Lag with strongest predictive importance \\
\midrule
\multicolumn{3}{@{}l}{\textbf{Causal discovery}} \\
\url{method/<m>/1-raw/results_<m>.csv} & 3 & Edges detected per method, with lags \\
\url{figures/diagnostics/<m>_fdr} & 3 & Edges surviving FDR correction per method \\
\url{figures/diagnostics/<m>_dag} & 3 & Consistency of directed edges per method \\
\url{figures/per_method/<m>_graph.svg} & 3 & Recovered graph per method \\
\url{figures/per_method/<m>_pvalues.svg} & 3 & Significance level per edge per method \\
\url{figures/per_method/<m>_lags.svg} & 3 & Detected effect lags per method \\
\url{figures/comparison/method_comparison.svg} & 3 & Edges found by multiple methods \\
\url{figures/comparison/pvalue_comparison.svg} & 3 & Significance comparison across methods \\
\url{figures/diagnostics/lag_analysis} & 3 & Cross-method lag analysis and agreement \\
\midrule
\multicolumn{3}{@{}l}{\textbf{Consensus}} \\
\url{ensemble_edges.csv} & 5 & Edges agreed by multiple methods \\
\url{consensus/2-core/consensus.csv} & 5 & $\ge\!2$-method agreement before tiering \\
\midrule
\multicolumn{3}{@{}l}{\textbf{Surrogate diagnostics and consensus support}} \\
\url{consensus/5-tiers/consensus_with_tiers.csv} & 5 & Method-count consensus with support-tier labels \\
\url{falsification_results.csv} & 6 & Dataset-level IAAFT surrogate edge-rate results \\
\multicolumn{3}{@{}l}{\textbf{Reproducibility and evaluation}} \\
\url{experiment_log.json} & 7 & Executed configuration (reproducibility record) \\
\url{graph_recovery_metrics.csv} & 7 & Method performance against ground truth \\
\bottomrule
\end{tabular*}
\end{table*}
\renewcommand{\arraystretch}{1}

\paragraph{Usage.}

Listings~\ref{lst:usage_default}--\ref{lst:usage_advanced} illustrate three levels of interaction with the software. The first executes the complete workflow using the recorded diagnostic and implementation defaults. The second restricts the run to pre-discovery diagnostics and non-causal reference models, which supports an initial inspection of the data before causal discovery. The third supplies an explicit configuration in which the selected methods, conditional-independence test, and $\tau_{\max}$ are informed by both the diagnostic output and domain knowledge.

\begin{lstlisting}[style=python,caption={Minimum invocation of AutoCause. The complete workflow runs using the recorded diagnostic and implementation defaults, and the resulting configuration is stored in the output log.},label={lst:usage_default}]
import pandas as pd
from framework.core.run_workflow import run_causal_discovery_workflow

df = pd.read_csv("my_data.csv", index_col=0, parse_dates=True)
result = run_causal_discovery_workflow(
    data_df=df,
    output_dir="results/my_experiment",
)
print(result["consensus"]["tier1_edges"])  # majority-supported candidate links
\end{lstlisting}

\begin{lstlisting}[style=python,caption={Exploratory invocation. The workflow first reports the pre-discovery diagnostics and then runs the lagged-correlation and predictive reference models. These outputs describe the associative and predictive structure of the data before any causal-discovery method is selected.},label={lst:usage_audit}]
result = run_causal_discovery_workflow(
    data_df=df,
    output_dir="results/exploration",
    enable_causal_audit=True,      # run pre-discovery diagnostics
    method_config={
        "granger":             {"enabled": False},
        "transfer_entropy":    {"enabled": False},
        "pcmci":               {"enabled": False},
        "varlingam":           {"enabled": False},
        "lpcmci":              {"enabled": False},
        "correlation":         {"enabled": True},   # association baseline
        "predictive_baseline": {"enabled": True},   # RF predictive baseline
    },
)
# Outputs: pre-discovery figures + correlation heatmap + RF feature importances
\end{lstlisting}

\begin{lstlisting}[style=python,caption={Full discovery run configured after the exploratory step in Listing~\ref{lst:usage_audit}. VARLiNGAM is enabled because the diagnostics indicate non-Gaussianity, whereas LPCMCI is excluded because of its computational cost. The practitioner supplies $\tau_{\max}$ using domain knowledge},label={lst:usage_advanced}]
result = run_causal_discovery_workflow(
    data_df=df,
    output_dir="results/discovery",
    tau_max=5,                     # from domain knowledge (e.g., 30h at 6h resolution)
    alpha=0.05,
    sampling_days=0.25,            # 6-hourly data = 0.25 days
    method_config={
        # Enable methods whose assumptions match the data (from Listing 2):
        "granger":             {"enabled": True},
        "transfer_entropy":    {"enabled": True},
        "pcmci":               {"enabled": True,
                                "test_method": "parcorr",  # diagnostics favored a linear test
                                "allow_contemporaneous": True},
        "varlingam":           {"enabled": True},   # non-Gaussian noise detected
        "lpcmci":              {"enabled": False},  # enable on HPC if confounders suspected
        "predictive_baseline": {"enabled": True},
        "correlation":         {"enabled": True},
    },
    enable_consensus=True,         # aggregate into consensus-support tiers
    enable_causal_audit=True,      # log assumption risks for reproducibility
    true_edges={("X1", "Y"), ("X2", "Y")},  # optional: if ground truth available
    undirected_eval=True,
)
\end{lstlisting}

The framework includes an examples folder containing the three benchmark scripts, a high-performance computing (HPC) submission template, and the scripts to reproduce the experiments in this paper. The release corresponding to this paper is version \texttt{0.2.0}. Every numerical claim corresponds to a file in the released experiment results.

\subsection{Methodological requirements and how AutoCause addresses them}\label{sec:capabilities}

The interpretation of causal-discovery output depends on identifying assumptions and finite-sample requirements that cannot be verified completely from observational data. AutoCause records diagnostic evidence for six decision areas, summarised in Table~\ref{tbl:capabilities}, and reports unresolved risks rather than certifying validity.

\begin{table}[pos=t]
\caption{Six methodological requirements that AutoCause screens before and during discovery. None of the diagnostics listed in the right column can verify causal validity on its own. When domain knowledge of the system is available it should override any data-driven estimate.}\label{tbl:capabilities}
\small
\begin{tabular*}{\tblwidth}{@{}p{2.4cm}p{5.9cm}p{7.0cm}@{}}
\toprule
Requirement & Theoretical basis & AutoCause assessment \\
\midrule
Causal search horizon ($\tau_{\max}$) & Bounds the tested delays and should reflect the sampling design and plausible process timescales \citep{runge2023causal} & Estimated from autocorrelation-function (ACF) decay when no physical value is supplied, then bounded by a user limit and an effective-sample-size constraint (\texttt{tau\_max}) \\
Conditional independence test & Test power depends on functional form; linear tests have low power when a dependency has little linear component \citep{runge2018conditional} & Assessed via RESET, distance correlation, and Shapiro--Wilk diagnostics; selects ParCorr, RobustParCorr, or CMIknn accordingly \\
Method identifiability & Each method assumes specific data properties (linearity, non-Gaussianity, causal sufficiency) & Six assumption-violation risks quantified by causal-audit; maps to the method whose conditions are best satisfied \\
Sample size adequacy & Finite-sample power decreases as the conditioning problem grows \citep{runge2018conditional} & Per-method $T_{\min}$ computed from the variable count $N$, $\tau_{\max}$, and the missing fraction; warns or falls back when insufficient \\
Multiple-testing correction & Testing many candidate links raises the expected false-discovery proportion \citep{benjamini1995controlling} & Benjamini--Hochberg correction applied when a method returns comparable per-link or per-lag $p$-values; uncorrected outputs are labelled (\texttt{fdr\_method}) \\
Multi-method corroboration & Repeated detection can prioritize links but does not establish causality \citep{assaad2022survey} & Method-count consensus assigns a support tier; dataset-level surrogate diagnostics are reported separately \\
\bottomrule
\end{tabular*}
\end{table}

\paragraph{Method selection.} Causal-audit maps the six risk scores to a recommended method and records the rule path that produced the recommendation. The recommendation is advisory because observational diagnostics cannot verify every identifying assumption, and users can replace it with a domain-informed method set. The full benchmark loops run four causal methods to measure complementarity. These results should therefore not be interpreted as an independent validation of the deployed selector.

\paragraph{Lag-window selection.} Each causal-discovery method requires a maximum lag, $\tau_{\max}$, that limits how far into the past the search for candidate causes extends. A window that is too short excludes delayed effects occurring beyond the selected horizon. A window that is too long increases both the number of tested links and the size of the conditioning sets, which can reduce statistical power. AutoCause therefore allows the user to define $\tau_{\max}$ from process knowledge whenever such information is available. For river discharge sampled every 6 hours, for example, an analyst may choose $\tau_{\max}=5$ when a 30-hour horizon exceeds the expected flood-wave travel time between connected gauges. A study of soil-moisture responses at daily resolution may require a longer lag window. When no physically supported value is available, AutoCause derives two fallback estimates: the first zero crossing of the autocorrelation function \citep{box2015time}, and the largest lag that preserves at least five effective observations per estimated coefficient \citep{lutkepohl2005var}. The smaller estimate is retained, subject to any upper bound supplied by the user. The run log stores the candidate values, the active constraint, and the final lag, allowing the analysis to be repeated with an alternative setting. In the experiments reported here, $\tau_{\max}=5$ is fixed for all methods so that lag selection does not contribute additional variation to the benchmark comparison.

\paragraph{Adaptive conditional-independence test.}
The conditional-independence test used by PCMCI+ directly affects which forms of dependence can be detected. ParCorr is computationally efficient and is appropriate for approximately linear conditional relationships. It can, however, fail when the association is strongly nonlinear, as in symmetric quadratic relationships for which linear correlation may vanish. CMIknn estimates conditional mutual information with a $k$-nearest-neighbour procedure and can represent a broader range of functional relationships, but its computational cost and sample-size requirements are substantially higher \citep{runge2018conditional}. AutoCause selects among the available tests using two pairwise diagnostics applied after preprocessing. The first is the Ramsey RESET test \citep{ramsey1969tests}, which fits a linear model of each variable from its own lags and tests whether polynomial functions of the fitted values explain additional residual variation. A significant result indicates that the linear specification may be inadequate. The second diagnostic compares distance correlation \citep{szekely2007measuring} with Pearson correlation for each variable pair. A comparatively large distance-correlation value indicates dependence that is not well represented by a linear measure. If either diagnostic is triggered, PCMCI+ is configured with CMIknn. Otherwise, a Shapiro--Wilk test on the residual distributions determines whether the workflow uses ParCorr or RobustParCorr, a rank-based alternative intended for departures from Gaussianity. The run log records the selected conditional-independence test, the diagnostic statistics, the corresponding $p$-values, and the rule that determined the selection. Because RESET and distance correlation assess marginal or pairwise relationships, they may not detect nonlinear dependence that appears only after conditioning on other variables. The user may therefore override the automatic selector and configure PCMCI+ to use CMIknn directly. The CI-sensitivity analysis in Table~\ref{tbl:edge_diagnostics} complements this option by rerunning PCMCI+ with ParCorr, RobustParCorr, and CMIknn, allowing the analyst to identify links whose detection changes with the selected test.

\paragraph{Sample-size screening and fallback.}
A nonparametric conditional-independence test may still return a graph when the available sample is too small for stable estimation. AutoCause therefore assesses sample-size adequacy before running causal discovery. It estimates the effective sample size, $T_{\text{eff}}$, after accounting for the selected lag window and missing observations, and compares the result with method-specific operational thresholds. The default values are 50 for ParCorr, 200 for CMIknn, 75 for VAR-Granger, and 120 for VARLiNGAM, with further adjustment for the number of variables and $\tau_{\max}$. These thresholds serve as implementation safeguards informed by published guidance and observed estimator behavior. They should not be interpreted as universal lower bounds or as guarantees of valid inference. When $T_{\text{eff}}$ falls below the threshold for the selected test, the workflow issues a warning and may use a less sample-intensive alternative, such as replacing CMIknn with ParCorr. The run log stores the applicable threshold, the estimated $T_{\text{eff}}$, and any fallback action, allowing reviewers and users to assess whether the reported analysis was supported by an adequate sample.

\paragraph{FDR handling.}
AutoCause applies Benjamini--Hochberg correction when a wrapped method returns comparable $p$-values over a clearly defined family of hypotheses. For VAR-Granger, the correction is applied across the candidate variable pairs evaluated at the selected lag. PCMCI+ uses the multiple-testing procedure implemented within its discovery rule. For transfer entropy, the correction covers the set of candidate lags tested for each variable pair. VARLiNGAM and the random-forest reference model do not return comparable per-edge $p$-values, so their outputs are reported without Benjamini--Hochberg adjustment. The corresponding result files explicitly indicate that no FDR correction was available. This design does not impose a common statistical interpretation on methods that expose different inferential quantities. Instead, AutoCause applies multiple-testing correction only when a method returns statistics that define a coherent family of hypotheses. The scope of each correction is stored in the run metadata.

\paragraph{Multi-method consensus and support tiering.} 
The benchmark evaluation includes four causal-discovery methods drawn from three methodological families: VAR-Granger and VARLiNGAM, transfer entropy, and PCMCI+. LPCMCI is available in the software but is omitted from the full benchmark loops because of its computational cost. Each discovered link is assigned a support tier according to the number of evaluated methods that detect it, using the fixed criteria in Table~\ref{tbl:tier_definitions}. Under this four-method composition, support from any three methods necessarily spans at least two methodological families so the family-diversity condition therefore does not impose an additional restriction beyond the majority threshold. The reported analysis consequently evaluates consensus by method count but does not determine whether agreement across different methodological families is more informative than agreement within the same family when the number of supporting methods is held constant.

\begin{table}[pos=t]
\caption{Consensus-support tiers used in the evaluation. These labels summarize agreement among the four evaluated causal-discovery methods and do not validate causality.}\label{tbl:tier_definitions}
\small
\begin{tabular*}{\tblwidth}{@{}p{1.1cm}p{6.2cm}p{7.0cm}@{}}
\toprule
Tier & Criterion & Interpretation \\
\midrule
Tier-1 & Detected by at least three of the four causal methods. With the present method set, this necessarily spans at least two methodological families. & High consensus support. The link is repeatedly detected under different modelling assumptions, but may still reflect indirect dependence or shared forcing. \\
Tier-2 & Detected by exactly two causal methods. & Moderate consensus support. The link has limited corroboration and requires method-specific and domain-based inspection. \\
Tier-3 & Detected by one causal method only. & Single-method support. The link is sensitive to one method's assumptions and should be treated as exploratory. \\
\bottomrule
\end{tabular*}

\vspace{4pt}
{\footnotesize Evaluated causal methods: VAR-Granger and VARLiNGAM (regression-based), transfer entropy (information-theoretic), and PCMCI+ (constraint-based). LPCMCI is available in the software but excluded from the full benchmark loops because of computational cost. Lagged correlation and the random-forest model are non-causal references. Dataset-level surrogate diagnostics are reported separately in Appendix~\ref{app:surrogate_falsification}.}
\end{table}

\paragraph{Per-edge diagnostic outputs.} In addition to the inferred graph and consensus-tier edge list, AutoCause generates the diagnostics listed in Table~\ref{tbl:edge_diagnostics}, which support interpretation when no reference graph is available. These outputs address whether an empty graph may reflect insufficient power, whether a reported link changes with the CI test, and whether its estimated strength varies across the record.

\begin{table}[pos=t]
\caption{Per-edge diagnostic outputs available to the practitioner after a discovery run. Each addresses a specific interpretability question that arises when ground truth is unavailable. MDES denotes the minimum detectable effect size.}\label{tbl:edge_diagnostics}
\small
\begin{tabular*}{\tblwidth}{@{}p{2.6cm}p{5.7cm}p{7.3cm}@{}}
\toprule
Diagnostic & What it reports & \textbf{Practitioner question answered} \\
\midrule
Power analysis (MDES) & Minimum detectable partial correlation given $T_\text{eff}$, $N$, $\tau_{\max}$ & ``No edges found: insufficient power, or genuine independence?'' \\
CI-test sensitivity & Edges found by ParCorr, RobustParCorr, and CMIknn separately & ``Does this edge survive a change of conditional-independence test?'' \\
Method-specific link strength & Normalized statistic per edge (partial $R^2$, normalized transfer entropy (TE), partial correlation) & ``How does this method's link statistic rank within the graph?'' \\
Temporal stability & Rolling-window statistic across temporal blocks & ``Does the reported link remain similar across the record?'' \\
Lag confidence & Bootstrap CI on the estimated lag per edge & ``What is the uncertainty range on the estimated lag?'' \\
\bottomrule
\end{tabular*}
\end{table}

\subsection{Pre-discovery diagnostics: extending causal-audit}\label{sec:predisc}

Causal-audit \citep{ruiz2026causalaudit} is a pre-discovery module that operationalizes the questionnaire-based method selector of \citet{runge2023causal} through statistical diagnostics. It covers branches that can be assessed from one stochastic time series without interventional data: stationarity, possible causal insufficiency, contemporaneous dependence, and restricted structural-model conditions. A high causal-insufficiency score can route the recommendation to LPCMCI, but it does not identify a hidden variable. Methods that require multiple related datasets, such as J-PCMCI+ and seqICP, approaches designed for deterministic dynamics, such as convergent cross mapping (CCM), and continuous-optimization methods, such as DYNOTEARS, are not included in the current implementation. AutoCause summarizes six data characteristics as risk scores between 0 and 1: nonstationarity, irregular sampling, persistence, possible causal insufficiency, nonlinearity, and seasonality. Each score is obtained through a logistic mapping and is accompanied by a bootstrap uncertainty interval. A rule-based decision tree then uses these scores to recommend a discovery method and conditional-independence test. When any prespecified risk threshold is exceeded, the workflow withholds an automatic recommendation and reports an abstention instead.

This paper extends the original causal-audit with three changes that emerged during DGP-Atlas validation. First, when all six risks for nonlinearity, confounding, nonstationarity, and irregularity score below their low-risk thresholds and the Shapiro--Wilk test rejects Gaussianity, the decision tree now recommends VARLiNGAM, whose ICA step can orient edges that ParCorr alone leaves ambiguous. Second, a defect in the diagnostic-extraction code was causing the nonlinearity and seasonality risk scores to return their logistic prior intercept regardless of the actual spectral-ratio and Spearman-minus-Pearson divergence observed in the data; correcting this defect restores the routing that directs seasonal series toward deseasonalization and nonlinear series toward CMIknn. Third, the original abstention rule rejected every dataset whose seasonality risk exceeded 0.60, even when deseasonalization followed by PCMCI+ is well-established practice \citep{box2015time,runge2020discovering}; the revised rule instead records a deseasonalization recommendation and proceeds. Because all three changes were developed with DGP-Atlas in view, Appendix~\ref{app:identifiability} treats the DGP-Atlas recommendation accuracy as an internal consistency check rather than as independent external validation.

\begin{figure}[pos=t]
  \centering
  \includegraphics[width=\textwidth]{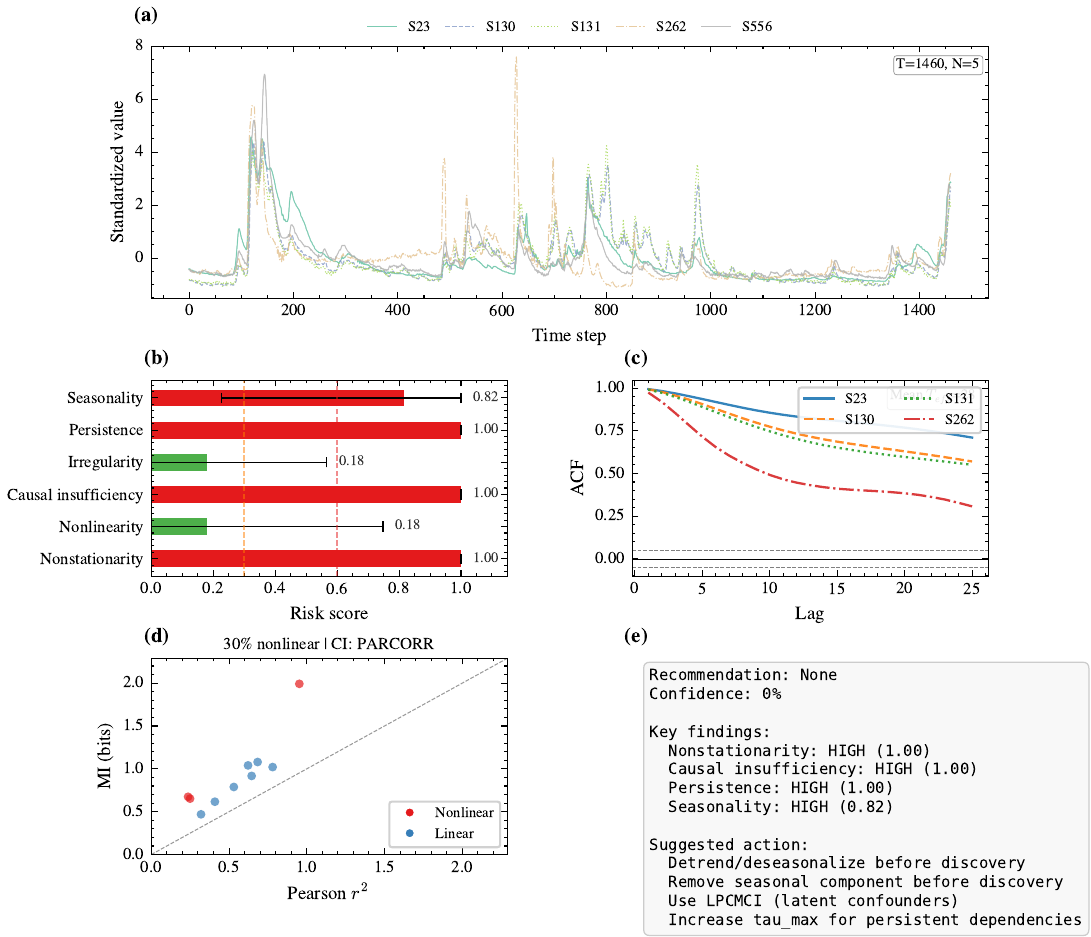}
      \caption{Pre-discovery summary for a CausalRivers subgraph comprising five Bavarian gauging stations on the Donau and Abens. The panels show the standardized discharge series; six assumption-violation risk scores with 95\% bootstrap uncertainty intervals, grouped as low ($<0.30$), moderate (0.30--0.60), and high ($\geq0.60$); variable-specific autocorrelation functions with 95\% significance bands and inset estimates of $T_{\text{eff}}$; pairwise nonlinearity diagnostics; and the resulting method and preprocessing recommendation. In the final panel, \texttt{None} denotes abstention because at least one prespecified high-risk threshold was exceeded.}\label{fig:prediscovery_example}
\end{figure}

Figure~\ref{fig:prediscovery_example} presents the pre-discovery summary generated for each dataset. It combines the estimated risk scores, autocorrelation profile, nonlinearity diagnostics, and recommended method in a single inspection panel. This summary supports review of the automated configuration, but it does not replace domain knowledge about the sampling process, physically plausible delays, or potentially omitted drivers.

\section{Experimental evaluation}\label{sec:experiments}

The evaluation uses the three benchmark collections introduced above: DGP-Atlas, TimeGraph, and CausalRivers, addressing these three questions:

\begin{itemize}
    \item How does the automated workflow perform relative to fixed or published configurations on the three benchmark collections?
    \item Do the evaluated causal-discovery methods recover complementary parts of the reference graph across datasets? 
    \item Does majority support identify links with higher precision, and under which environmental conditions does this ordering break down?
\end{itemize}

Reference-graph information is not provided to AutoCause during an individual run. It is introduced only after discovery to compute evaluation metrics. DGP-Atlas informed development of the pre-discovery module, while TimeGraph influenced the design of the benchmark comparison. The experiments therefore assess the software implementation in terms of graph recovery, complementarity among methods, precision across support tiers, and dataset-level surrogate behavior. They do not independently validate the automatic lag estimator, every sample-size fallback rule, or the thresholds used by the causal-audit module. The reported findings should consequently be interpreted as benchmark evidence for the evaluated release, not as external validation of all six automated decisions.

\subsection{Setup, dataset selection, and evaluation metrics}\label{sec:benchmarks}

The three benchmarks examine different aspects of the workflow. DGP-Atlas represents controlled violations of statistical assumptions, TimeGraph provides structured synthetic failure modes, and CausalRivers supplies environmental time series evaluated against a topology-derived river reference. For DGP-Atlas and TimeGraph, the known generative graph is used as ground truth. In CausalRivers, the directed river topology serves as an external reference, but it does not constitute a complete model of discharge dynamics. Shared rainfall, snowmelt, reservoir regulation, tributary inflows, and other omitted processes may produce dependencies that are absent from the topology. The primary evaluation is performed on graph skeletons so each reported link is treated as an undirected adjacency between its two endpoint variables, discarding its orientation mark and its estimated lag. With TP, FP, and FN denoting the true-positive, false-positive, and false-negative adjacency counts, skeleton precision is $\mathrm{Precision} = \frac{\mathrm{TP}}{\mathrm{TP}+\mathrm{FP}}$. Recall, reported as the true-positive rate is $\mathrm{TPR} = \frac{\mathrm{TP}}{\mathrm{TP}+\mathrm{FN}}$. The F1 score is the harmonic mean of the two, $\mathrm{F1} = \frac{2\,\mathrm{Precision}\,\mathrm{TPR}}{\mathrm{Precision}+\mathrm{TPR}}$. Thus, the structural Hamming distance is $\mathrm{SHD} = \mathrm{FP}+\mathrm{FN}$, and the graph-level false discovery rate is $\mathrm{FDR} = \frac{\mathrm{FP}}{\mathrm{TP}+\mathrm{FP}}$. This quantity is the realized proportion of false links in the predicted graph. It differs from the expected false discovery rate targeted by the Benjamini--Hochberg multiple-testing procedure. When a method predicts no links, the released evaluation records graph FDR as zero. This convention must be interpreted together with TPR, because the precision denominator is empty in such cases.

On the synthetic benchmarks, FP and FN are measured against the known generating graph. In contrast, on CausalRivers, they are measured against direct upstream-to-downstream adjacencies in the topology-derived reference. The area under the receiver operating characteristic (AUROC) curve and the area under the precision--recall (AUPRC) curve are reported only for methods that supply a continuous score comparable across candidate links. This practically translates into using PCMCI+ (gives calibrated $p$-values for each individual edge), VAR-Granger (F-test $p$-values), and transfer entropy (permute-test based $p$-values). In the case of VARLiNGAM, we have estimates of connection strength but no associated $p$-values; in addition, random forest (baseline) gives us permutation importance values which do not make sense across pairs.

Furthermore, there are distinct identifiability conditions for lagged and contemporaneous relationships, hence the orientation problem is considered independently of the skeleton construction. While TimeGraph has both types of relationships and CausalRivers defines upstream-to-downstream oriented contemporaneous relationships, the algorithms have a different capability to orient the latter relationships. Thus, the skeleton reconstruction procedure becomes the common objective measure for all three benchmark sets, and the orientation issue is considered separately in Appendix~\ref{app:identifiability}.

\paragraph{Parameters and design choices.}

The main 145-dataset evaluation uses $\alpha=0.05$ and $\tau_{\max}=5$. Benjamini--Hochberg correction is applied wherever per-edge $p$-values are available: across candidate pairs for VAR-Granger, within the PCMCI+ procedure, and across candidate lags for each transfer-entropy pair. The lagged-correlation reference instead selects the smallest $p$-value across lags and two correlation tests and compares it directly with $\alpha=0.05$, without an additional correction across pairs. It is therefore the least conservative reference and is used only as an association screen. VARLiNGAM and the random-forest reference do not return per-edge $p$-values and are reported without this correction. No benchmark-specific hyperparameter search was conducted; the remaining choices follow from the pre-discovery diagnostics or stated operational constraints. Five further design choices are fixed across experiments: 

First, $\tau_{\max}=5$ is held fixed across all three benchmarks rather than re-estimated per dataset. The framework's automatic estimator (Section~\ref{sec:capabilities}) is bypassed so that every method searches the same lag window. The chosen value covers the true lag of two on TimeGraph with margin and matches the estimator's recommendation on a representative DGP-Atlas pre-discovery audit.

Second, the adaptive conditional-independence test selector operates through the pre-discovery module. The evaluated release uses fixed operational thresholds for NonlinearityRisk, SeasonalityRisk, and IrregularityRisk to determine the conditional-independence test and any preprocessing action. Because DGP-Atlas informed the development of this module, the corresponding thresholds are implementation defaults rather than universally validated cutoffs. AutoCause stores their values in the run configuration, allowing users to replace them when domain knowledge or a sensitivity analysis justifies an alternative setting.

Third, the DGP-Atlas configuration includes ten dataset identifiers per family. The pre-discovery module flags the high-persistence family for additional stationarity assessment, but those datasets remain in the stress-test evaluation. Some irregular-sampling datasets also fall below the operational effective-sample threshold; they are retained so that the reported results do not depend on the screening stage accepting the dataset. Three nonlinear datasets are excluded because their explosive trajectories violate the bounded-variance rule implemented in the release, leaving 97 DGP-Atlas datasets for evaluation.

Fourth, LPCMCI is excluded from the benchmark loops because the stated trial configuration exceeded the two-hour wall-time limit. Fifth, we resample CausalRivers from 15-minute to 6-hour resolution and restrict the analysis to calendar year 2021, which keeps runtime tractable while providing approximately 1460 observations per station. The lag horizon is fixed at $\tau_{\max}=5$, corresponding to a 30-hour search window that exceeds the flood-wave travel time between directly connected Alpine-foreland stations. This value is an operational benchmark setting chosen for cross-method comparability and is not claimed to cover every hydrological delay in the network. From each of the three topology classes described in Section~\ref{sec:benchmarks_review} we draw 10 subgraphs of five stations, giving 30 subgraphs total.

\subsection{Causal discovery results}\label{sec:results}

\begin{figure}[pos=t]
  \centering
  \includegraphics[width=\textwidth]{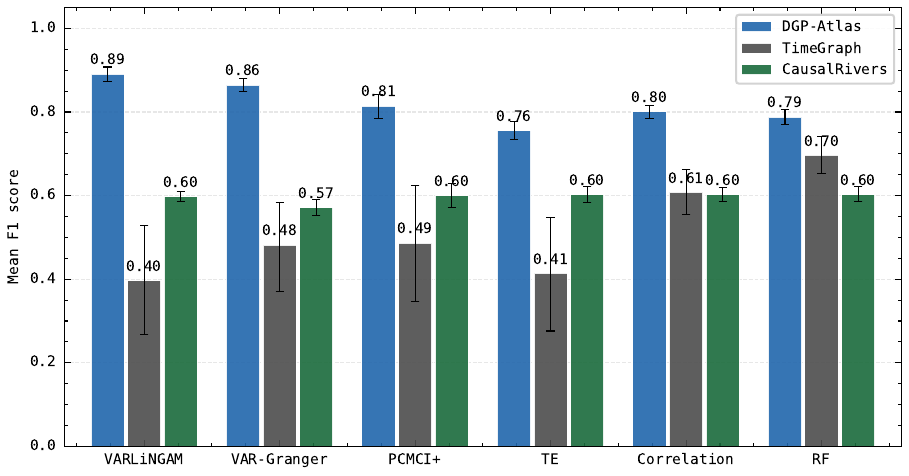}
  \caption{Mean F1 score per method per benchmark with dataset-level bootstrap 95\% intervals from 2000 resamples. Lagged correlation and the random-forest predictive baseline are non-causal references. Source: \texttt{make\_figures.py} and the per-dataset metric files specified in Section~\ref{sec:benchmarks}.}\label{fig:results_summary}
\end{figure}

Method performance appears in Figure~\ref{fig:results_summary} and Table~\ref{tbl:results_summary}. VARLiNGAM has the highest aggregate DGP-Atlas F1. On TimeGraph, the evaluated causal methods have lower recall in several nonlinear categories. CausalRivers causal-method point estimates lie between 0.57 and 0.60 and their intervals overlap. No paired test is used to claim equality. The random-forest model measures lagged predictability and remains a non-causal reference.

\begin{table}[pos=t]
\caption{Mean F1 score per method per benchmark using unordered skeleton pairs. Bold marks the highest point estimate per benchmark, with ties bolded. Non-causal references are not ranked as causal methods. LPCMCI was excluded from the benchmark because the trial configuration exceeded the predefined wall-time limit.}\label{tbl:results_summary}
\small
\begin{tabular*}{\tblwidth}{@{}p{3.5cm}p{2.5cm}p{2.5cm}p{2.5cm}@{}}
\toprule
Method & DGP-Atlas & TimeGraph & CausalRivers \\
\midrule
\multicolumn{4}{@{}l}{\textit{Causal methods}} \\
VAR-Granger         & 0.864 & 0.482 & 0.571 \\
VARLiNGAM           & \textbf{0.889} & 0.396 & 0.597 \\
Transfer entropy    & 0.756 & 0.414 & \textbf{0.602} \\
PCMCI+ (adaptive CI) & 0.814 & 0.486 & 0.600 \\
\midrule
\multicolumn{4}{@{}l}{\textit{Non-causal baselines references}} \\
Lagged correlation  & 0.801 & 0.608 & \textbf{0.602} \\
Random Forest & 0.788 & \textbf{0.696} & \textbf{0.602} \\
\bottomrule
\end{tabular*}
\end{table}

\begin{figure}[pos=t]
  \centering
  \includegraphics[width=\textwidth]{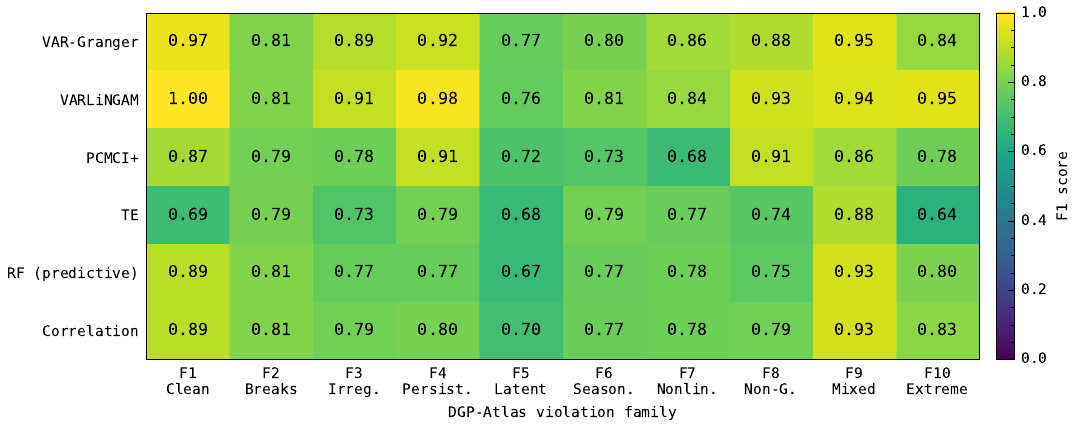}
  \caption{Mean F1 score for each method across the 10 DGP-Atlas families, including the lagged-correlation and predictive reference models. The preprocessing stage applies interpolation to F3 and F9 and removes seasonality from F6. VARLiNGAM achieves the highest mean F1 in six families, while the lagged-correlation baseline remains slightly below the best-performing causal method in each family.}\label{fig:atlas_heatmap}
\end{figure}

Because the 10 DGP-Atlas families represent different assumption violations, the aggregate F1 values in Table~\ref{tbl:results_summary} conceal substantial variation across regimes. The disaggregated result can be seen in Figure~\ref{fig:atlas_heatmap}, which reports the family-level results for the causal methods and the two non-causal reference models. VARLiNGAM obtains the highest mean F1 in six families: F1, F3, F4, F6, F8, and F10. Its clearest advantage occurs on F8, which introduces non-Gaussian noise, where it reaches $\mathrm{F1}=0.93$. This result is consistent with the ICA-based identifiability assumptions of LiNGAM \citep{hyvarinen2010estimation,shimizu2006lingam}. PCMCI+ reaches $\mathrm{F1}=0.87$ on the clean VAR family F1 and $\mathrm{F1}=0.91$ on F8.

Preprocessing has a marked effect on the irregular-sampling families. On F3, linear interpolation before discovery raises PCMCI+ from $\mathrm{F1}=0.00$ to $\mathrm{F1}=0.78$, while VARLiNGAM reaches $\mathrm{F1}=0.91$ under the same preprocessing. On F9, which combines irregular sampling with additional violations, PCMCI+ increases from $\mathrm{F1}=0.26$ without interpolation to $\mathrm{F1}=0.86$ after the audit-triggered step. The lagged-correlation baseline remains competitive, with F1 values between 0.70 and 0.93, but it does not lead any family. F4 provides a separate stress test for persistence: although the pre-discovery module flags the family for stationarity risk, both VARLiNGAM and PCMCI+ retain mean F1 scores above 0.90.

\begin{figure}[pos=t]
  \centering
  \includegraphics[width=\textwidth]{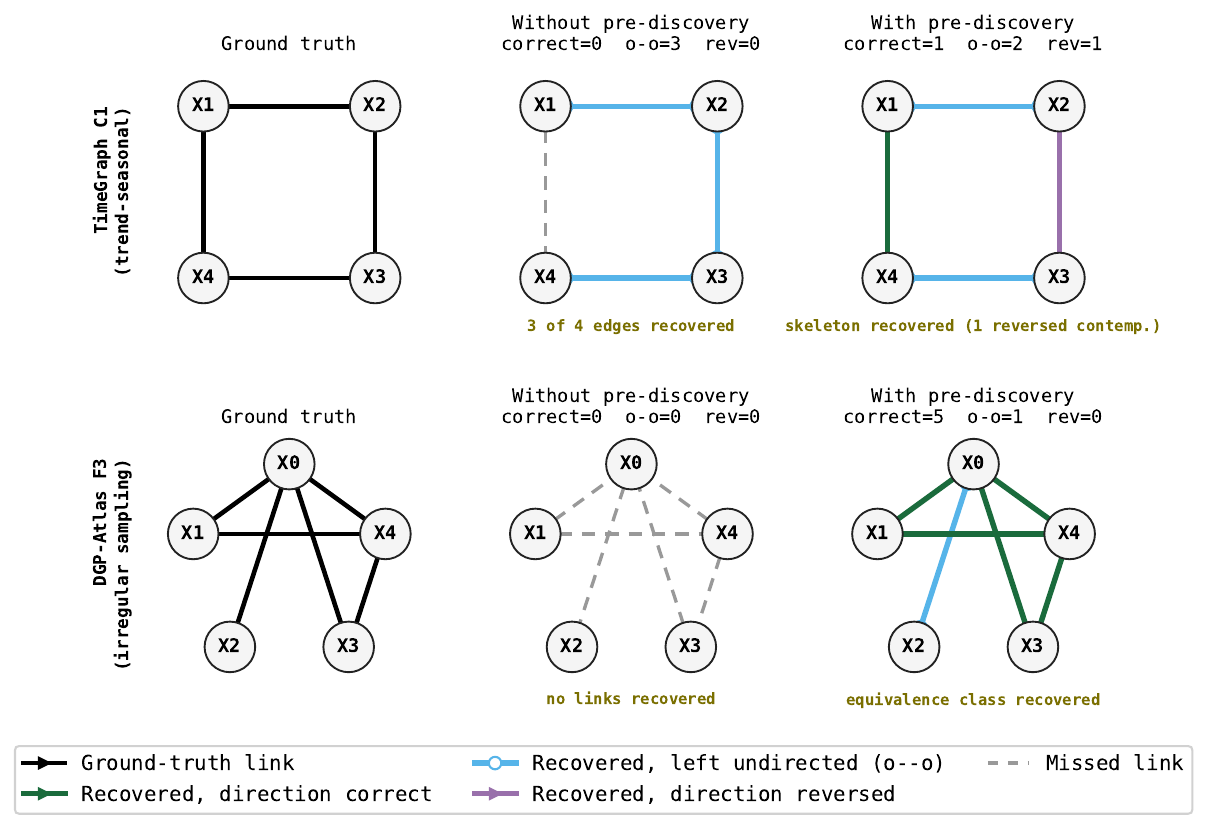}
  \caption{Effect of audit-driven preprocessing on two cases. Both rows run PCMCI+ (ParCorr, Benjamini--Hochberg FDR); the two right-most columns differ only in whether the audit-recommended preprocessing is applied. Top row, TimeGraph C1 (trend-seasonal): without deseasonalization three of four skeleton edges are recovered; with it all four are recovered, with one contemporaneous link oriented against the equivalence class. Bottom row, DGP-Atlas F3 (irregular sampling, 27\% missing): without imputation no edges are recovered; with audit-triggered interpolation the skeleton is exact (TP = 6, FP = 0) and all lagged links are oriented. Green: correct skeleton and direction. Blue \texttt{o\,--\,o}: contemporaneous link left undirected within the Markov equivalence class. Purple: recovered but oriented against the class. Grey dotted: missed. Orange: false positive.}\label{fig:skeleton_direction}
\end{figure}

\begin{figure}[pos=t]
  \centering
  \includegraphics[width=\textwidth]{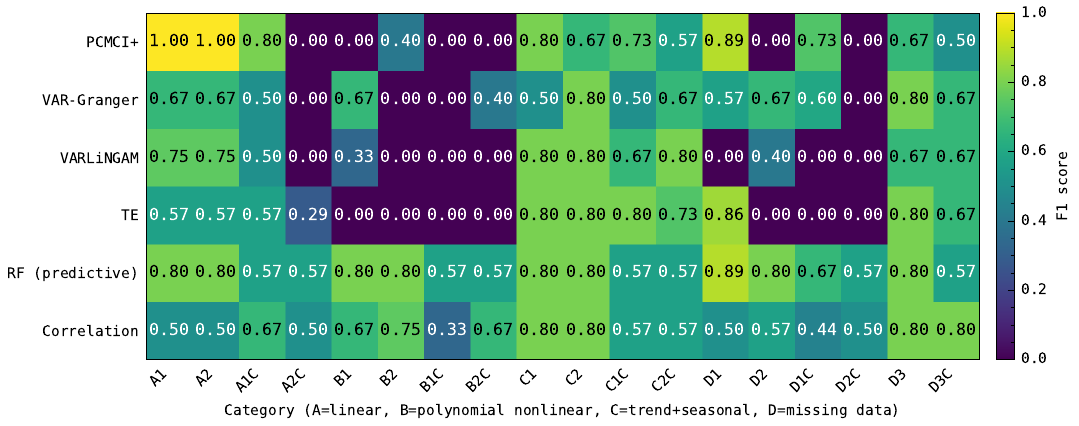}
  \caption{F1 score for each method across the 18 TimeGraph categories, evaluated on unordered skeleton pairs. Lagged correlation and the predictive model are included as non-causal reference methods. The figure was generated with \texttt{make\_figures.py} from the TimeGraph per-family result files.}
  \label{fig:timegraph_heatmap}
\end{figure}

\begin{figure}[pos=t]
  \centering
  \includegraphics[width=\textwidth]{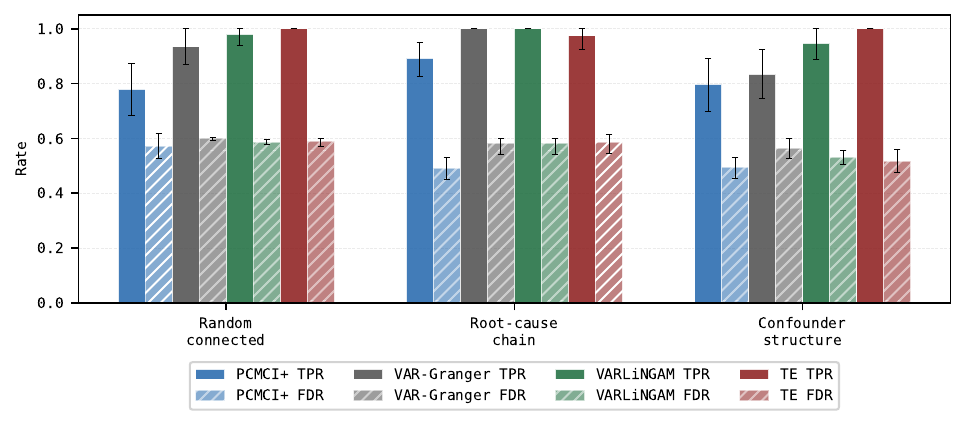}
  \caption{Skeleton TPR (solid) and graph FDR (hatched) for CausalRivers, grouped by subgraph topology, with bootstrap 95\% intervals across subgraphs. The three classes are the CausalRivers sampling strategies: randomly connected five-station subgraphs, root-cause chains in which the longest directed path contains all five stations, and confounder structures in which at least one station has two or more downstream stations. PCMCI+ has the lowest FDR point estimate on both the root-cause and confounder subgraphs, with 0.49 in each case.}
  \label{fig:causalrivers_fdr}
\end{figure}

TimeGraph and CausalRivers break this pattern into per-category and per-topology detail (Figures~\ref{fig:timegraph_heatmap} and~\ref{fig:causalrivers_fdr}). On the linear TimeGraph categories A1 and A2, PCMCI+ reaches F1 = 1.00 (TPR = 1.00, FDR = 0.00, SHD = 0), matching the published baseline. The polynomial-nonlinear B categories tell a different story: ParCorr has no power against cubic and quadratic dependence, so PCMCI+ and transfer entropy drop to near zero. VAR-Granger still detects the linear term embedded in the structural equations (B1, F1 $\approx$ 0.67). The correlation and predictive baselines hold F1 $\approx$ 0.67--0.80 by capturing marginal association without conditioning. Figure~\ref{fig:skeleton_direction} illustrates the preprocessing effect on one trend-seasonal TimeGraph case and one irregularly sampled DGP-Atlas case; Appendix~\ref{app:identifiability} provides the orientation analysis.

The CausalRivers study evaluates a different river network, subgraph sample, and tuning protocol \citep{stein2025causalrivers}. Its published values are therefore contextual rather than direct baselines for the fixed-configuration RiversBavaria evaluation reported here.

The topology-specific results (Figure~\ref{fig:causalrivers_fdr}) vary across methods and across the three subgraph classes defined in Section~\ref{sec:benchmarks_review}: randomly connected subgraphs, root-cause chains, and confounder subgraphs. PCMCI+ yields the lowest observed FDR on the evaluated root-cause and confounder subgraphs. This pattern is consistent with conditional-independence testing removing some indirect associations, although the aggregate metrics do not isolate that mechanism \citep{runge2020discovering}. Transfer entropy attains the highest F1 point estimate on the confounder subgraphs. Determining whether this advantage reflects nonlinear dependence induced by shared upstream drivers would require a path-level analysis, which is outside the present evaluation.

The connection between topologically adjacent stations creates strong lagged dependence in the discharge signal, explaining the high recall measured against the topology-derived reference. Predicted links outside that reference may arise from indirect propagation, unmeasured meteorological forcing, reservoir regulation, tributary inflows, or other catchment processes. Because these mechanisms were not evaluated directly, they should be treated as possible explanations rather than identified causes of the observed error pattern. Across the evaluated methods and topology classes, graph FDR ranges from 0.49 to 0.60.

All CausalRivers results reported here are based on 30 subgraphs of five stations, discharge records from calendar year 2021, and a 6-hour sampling interval. The sample represents a limited rainfall--nival regime in Bavaria and differs from the flatter East German network examined by \citet{stein2025causalrivers}. Regulated rivers, snowmelt-dominated catchments, or networks evaluated at different temporal resolutions may produce different method rankings. Similar performance across the three selected topology classes therefore does not establish transfer to other river systems, climates, lag windows, or monitoring designs.

\begin{figure}[pos=t]
  \centering
  \includegraphics[width=\textwidth]{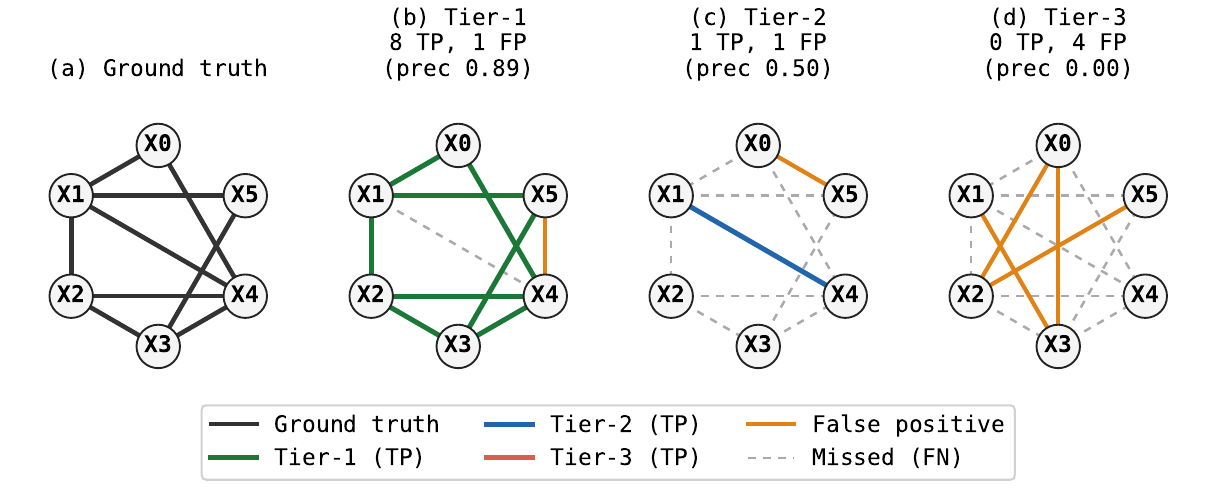}
  \caption{Method-count support for one DGP-Atlas instance from the non-Gaussian F8 family (\texttt{dgp\_002}), containing six variables and nine reference connections. All results are evaluated on the unordered skeleton. Panel (a) shows the reference graph. Panel (b) shows Tier-1 links detected by at least three methods, recovering 8 of 9 reference links with precision 0.89. Panel (c) shows Tier-2 links detected by two methods, of which 1 of 2 matches the reference. Panel (d) shows Tier-3 links detected by one method, none of which matches the reference. Orange edges are absent from the reference graph, and grey dashed edges denote missed reference connections. The figure was generated with \texttt{reproduce\_fig8\_tiers.py}.}
  \label{fig:causal_graph_example}
\end{figure}

The following analyses examine method complementarity (Section~\ref{sec:ablation}) and test whether majority support increases precision (Section~\ref{sec:tiering_validation}). Figure~\ref{fig:causal_graph_example} illustrates the support-tier rule on one DGP-Atlas example.

\subsection{Multi-method evaluation and comparison with reported results} 
\label{sec:ablation} 

The TimeGraph analysis compares the PCMCI+ results produced by AutoCause with the published values reported by \citet{ferdous2025timegraph} on the same datasets. The two evaluations differ in their preprocessing steps, graph-matching conventions, and multiple-testing procedures. The comparison therefore characterizes the behavior of the two pipelines rather than isolating the effect of any single design choice. It is a diagnostic benchmark, not a component ablation, and is separate from the 18-category main evaluation. Its ten rows run under a separate released configuration rather than the main-evaluation defaults of Section~\ref{sec:benchmarks}. Within this diagnostic run, the categories, sample length, variable count, lag setting, and ParCorr test are aligned; AutoCause retains its own input handling and multiple-testing procedure. SHD values are excluded because the two pipelines use incompatible graph conventions. Attributing any performance difference to a single stage would require an isolated component experiment.

\begin{table}[pos=t]
\caption{Published PCMCI+ values \citep{ferdous2025timegraph} and AutoCause PCMCI+ values on ten reported TimeGraph rows. Each column retains its pipeline's graph-matching convention; the comparison is diagnostic and does not estimate a component effect. FDR is recorded as zero for an empty prediction under the convention defined in Section~\ref{sec:benchmarks}.}\label{tbl:baseline_comparison}
\small
\begin{tabular*}{\tblwidth}{@{}p{4.2cm}p{1.7cm}cccc@{}}
\toprule
& & \multicolumn{2}{c}{Published PCMCI+} & \multicolumn{2}{c}{AutoCause PCMCI+} \\
\cmidrule(lr){3-4} \cmidrule(lr){5-6}
Category & Noise & TPR & FDR & TPR & FDR \\
\midrule
A1 (linear)                  & Gaussian   & \textbf{1.00} & \textbf{0.00} & \textbf{1.00} & \textbf{0.00} \\
A1 (linear)                  & Student-t  & 0.67 & 0.33 & \textbf{1.00} & \textbf{0.00} \\
B1 (poly. nonlinear)         & Gaussian   & \textbf{0.00} & 1.00 & \textbf{0.00} & \textbf{0.00} \\
B1 (poly. nonlinear)         & Student-t  & 0.33 & 0.93 & \textbf{1.00} & \textbf{0.33} \\
C1 (trend-seasonal)          & Gaussian   & 0.00 & 1.00 & \textbf{1.00} & \textbf{0.33} \\
A1C (linear + confounder)    & Gaussian   & 0.67 & 0.50 & \textbf{1.00} & \textbf{0.20} \\
A1C (linear + confounder)    & Student-t  & 0.67 & 0.50 & \textbf{1.00} & \textbf{0.00} \\
B1C (poly. nonlin. + conf.)  & Gaussian   & \textbf{0.22} & \textbf{0.00} & 0.00 & \textbf{0.00} \\
B1C (poly. nonlin. + conf.)  & Student-t  & 0.33 & 0.94 & \textbf{1.00} & \textbf{0.33} \\
C1C (trend-seas. + conf.)    & Gaussian   & 0.33 & 0.79 & \textbf{0.50} & \textbf{0.33} \\
\midrule
Mean                          &            & 0.42 & 0.60 & \textbf{0.75} & \textbf{0.15} \\
\bottomrule
\end{tabular*}
\end{table}

The pipelines agree on the clean A1 Gaussian row and differ on several seasonal, heavy-tailed, and confounded rows. Both have low TPR on at least one Gaussian polynomial-nonlinear row. Because graph conventions and processing pipelines differ, the table describes reported behavior rather than a matched effect estimate. Attributing a difference to one decision rule would require a component ablation.

Table~\ref{tbl:ablation} compares PCMCI+ with a per-dataset oracle that selects the highest-F1 result among the four evaluated causal methods and configurations. The oracle is not deployable because it uses the reference graph. The oracle achieves a higher F1 score than PCMCI+ on each of the three benchmark collections, while the identity of the best-performing method varies across datasets. This establishes complementarity within the evaluated set, not the performance of a deployable selector.

\begin{table}[pos=t]
\caption{PCMCI+ versus a per-dataset oracle that chooses the highest-F1 result among the four evaluated causal methods and configurations using the reference graph. The oracle bounds selectors restricted to this evaluated set; it is not deployable. The final column reports the most frequent sole winner. Skeleton F1 is used throughout.}\label{tbl:ablation}
\small
\begin{tabular*}{\tblwidth}{@{}lcccl@{}}
\toprule
Benchmark    & PCMCI+ alone & Multi-method oracle & $\Delta$F1 & Most frequent best method \\
\midrule
DGP-Atlas    & 0.814 & \textbf{0.918} & $+0.104$ & VARLiNGAM (27/97) \\
TimeGraph    & 0.486 & \textbf{0.639} & $+0.153$ & PCMCI+ (6/18) \\
CausalRivers & 0.600 & \textbf{0.648} & $+0.048$ & PCMCI+ (14/30) \\
\bottomrule
\end{tabular*}
\end{table}

A separate question concerns the vote threshold used to turn those multi-method outputs into a prioritized edge list. Table~\ref{tbl:consensus_ablation} compares the majority-of-three rule used for Tier-1 with looser and stricter vote thresholds. In the current four-method set, the majority-of-three and the stated family-aware rule are mathematically identical because any three methods necessarily span at least two methodological families. The reported precision gain therefore comes from vote count, not from an independently demonstrated family-diversity effect. Requiring any two methods admits more links at lower precision, while unanimity admits fewer links at higher precision. Whether cross-family agreement carries information beyond the vote count alone requires an equal-vote comparison between same-family and cross-family pairs, not attempted here.

\begin{table}[pos=t]
\caption{Vote-threshold precision on DGP-Atlas and TimeGraph. Rules admit a link by agreement among the four causal methods: majority-3 (the Tier-1 rule), any-two, and unanimous. The family-aware and majority-3 rules are identical for the present method set. $n$ is the number of admitted edges, so stricter rules admit fewer at higher precision. Bold marks the majority-of-three rule used for Tier-1.}\label{tbl:consensus_ablation}
\small
\begin{tabular*}{\tblwidth}{@{}lcccc@{}}
\toprule
Rule & DGP-Atlas prec. & DGP-Atlas $n$ & TimeGraph prec. & TimeGraph $n$ \\
\midrule
Family-aware majority ($\geq$3 methods) & \textbf{0.848} & 1291 & \textbf{0.694} & 49 \\
Majority-3 ($\geq$3 methods) & 0.848 & 1291 & 0.694 & 49 \\
Any two ($\geq$2 methods) & 0.778 & 1500 & 0.562 & 80 \\
Unanimous (all 4 causal methods) & 0.905 & 877 & 0.706 & 17 \\
\bottomrule
\end{tabular*}
\end{table}

\subsection{Consensus-support precision evaluation}\label{sec:tiering_validation}

Figure~\ref{fig:causal_graph_example} illustrates method-count consensus on one DGP-Atlas instance. Across DGP-Atlas, Tier-1 precision is 0.848, compared with 0.344 for Tier-2 and 0.126 for Tier-3. TimeGraph shows the same ordering: 0.694, 0.355, and 0.300. Majority support therefore prioritizes higher-precision links on the two synthetic benchmarks. These values are skeleton precision, not calibrated posterior probabilities or causal-validity scores.

The ordering does not hold on CausalRivers. Relative to the topology-derived reference, Tier-1 precision is 0.425, while Tier-2 and Tier-3 precision are 0.667 and 0.571. The latter values use six and seven links, respectively. The observed result is that majority agreement does not select the highest-precision tier in this benchmark. Omitted shared drivers are one possible explanation, but the evaluation does not identify their contribution.

\begin{figure}[pos=t]
  \centering
  \includegraphics[width=1.0\textwidth]{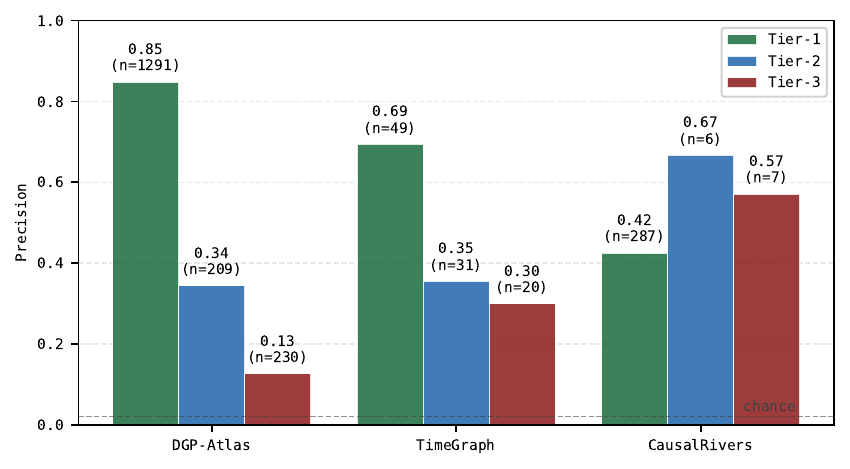}
  \caption{Consensus-support precision per benchmark. On DGP-Atlas and TimeGraph, majority-supported Tier-1 exceeds Tier-2 and Tier-3. On CausalRivers the ordering reverses; the Tier-2 and Tier-3 estimates use only six and seven edges. The dashed line shows a common reference prevalence. Sources: \texttt{recompute\_tier\_metrics.py} and \texttt{verify\_causalrivers\_tiers.py}.}\label{fig:tiering}
\end{figure}

\section{Discussion}\label{sec:discussion}

AutoCause does not relax the assumptions required by time-series causal discovery or turn observational records into causal proof. It records assumption diagnostics, preprocessing, lag selection, conditional-independence testing, FDR handling, method comparison, surrogate checks, and support tiers. The analysis path is therefore inspectable even for environmental records with autocorrelation, seasonality, missing observations, nonlinear responses, or unmeasured common drivers. None of the recorded steps guarantees identification.

\subsection{From method selection to auditable causal-discovery workflow}

No causal method is the sole best result on every evaluated dataset, consistent with broader benchmark evidence \citep{runge2023causal,assaad2022survey}. PCMCI+ with ParCorr targets approximately linear conditional relationships, VARLiNGAM uses non-Gaussian linear structure, information-theoretic methods offer nonparametric tests at greater sample-size cost, and LPCMCI represents latent-confounding ambiguity with greater computational cost. The workflow records which assumptions and constraints informed each route rather than treating a common API as method selection.

Table~\ref{tbl:baseline_comparison} describes two processing pipelines rather than an isolated component effect. They agree on the clean linear row and differ elsewhere, but their graph conventions, preprocessing, input handling, and multiple-testing procedures prevent attribution to one decision rule. In nonlinear categories, an empty or unstable graph should be reported with the corresponding sample-size and test diagnostics rather than interpreted as causal absence.

AutoCause applies Benjamini--Hochberg correction when a method exposes a compatible testing family \citep{benjamini1995controlling}. The TimeGraph comparison does not isolate the contribution of Benjamini--Hochberg correction. It does, however, show why the correction procedure, the associated family of hypotheses, and the convention used for empty predictions must be reported explicitly.

\subsection{Consensus support and its limits}

The consensus-support scheme converts agreement across methods into an ordering for further inspection. On DGP-Atlas and TimeGraph, links supported by three or four methods have higher precision than links reported by only one method. These tiers summarize observed agreement across the evaluated methods; they are not posterior probabilities or guarantees of causal validity. 

CausalRivers shows that this ordering is not universal. All methods report links outside the topology-derived reference as well as reference adjacencies, so several methods can agree on a dependency that is absent from the direct topology, possibly because of indirect propagation or omitted common drivers.

The tier-precision inversion follows from this difference in reference structure. On the synthetic benchmarks, known graphs and controlled violations make majority support a useful precision filter. On river data, Tier-1 marks a dependency that needs physical interpretation through topology, travel time, elevation, regulation, tributaries, catchment boundaries, and measured meteorological inputs. Cross-method agreement does not remove latent-confounding or omitted-variable limitations \citep{peters2017elements, spirtes2000causation}.

For environmental modelling, the support tier should be read together with pre-discovery diagnostics and domain context. Elevated seasonality, persistence, or causal-insufficiency risk weakens the interpretation of any consensus-supported link. Tier-1 identifies a priority for further analysis, not a direct causal conclusion.

\subsection{Falsification and the role of non-causal baselines}

IAAFT surrogates add a dataset-level check beyond graph-recovery metrics. Each surrogate retains the marginal distribution and approximately preserves the power spectrum of every variable while disrupting dependence between variables. Methods that continue to report many links under this null produce a high surrogate edge rate. On the evaluated CausalRivers subset, PCMCI+ has a lower surrogate edge rate than the regression-based methods. This result characterizes aggregate behavior under the surrogate null and does not establish the validity of any individual observed link.

Lagged correlation and the random-forest baseline address different questions from the causal-discovery methods. The former summarizes temporal association, while the latter captures predictive structure. Neither provides evidence of causality by itself. A gap between predictive and causal recovery can be consistent with nonlinearity, omitted drivers, insufficient sample size, or violated identifiability assumptions, but it does not identify which explanation applies. The references therefore distinguish association and prediction from dependencies that survive a causal method's conditioning and correction procedure. The predictive baseline in particular requires careful reading. It is excluded from the causal voting rule and its significance flags are not corrected in the same way as the causal methods. Its advantage on some TimeGraph categories is a predictive reference point, not a directly comparable causal-discovery result. The asymmetry is diagnostically useful because it exposes structure that causal methods miss, but it should not be used to rank random forests as causal methods.

\subsection{Decisions that still require domain expertise}

Automation does not remove the need for domain judgment. Three decisions remain especially dependent on knowledge of the underlying system.

The first concerns the maximum lag, $\tau_{\max}$. An autocorrelation-based estimate provides an operational starting point, but it does not establish a physically meaningful delay horizon. Information about hydrological travel time, ecological response delay, sampling frequency, and known process memory should take precedence when available. A lag window that is too short excludes delayed effects beyond the selected horizon. A window that is too long increases the number of tested links, enlarges the conditioning sets, and can reduce statistical power.

The second concerns nonlinear conditional-independence testing. CMIknn can detect a broader range of functional relationships than linear tests, but its performance depends strongly on effective sample size and conditioning dimension. In the evaluated settings, some nonlinear regimes remain difficult to recover despite the use of a nonlinear test. An empty or sparse result from CMIknn should therefore be interpreted together with the corresponding sample-size and power diagnostics rather than as evidence that no causal dependence is present.

The third concerns latent confounding. No software default can establish that all relevant common drivers have been observed. LPCMCI represents uncertainty caused by latent confounders, but its computational cost exceeded the wall-time budget used in the benchmark evaluation. When the pre-discovery diagnostics indicate a high risk of causal insufficiency, stronger analysis may require additional variables, explicit domain constraints, independent validation, interventional evidence, or multiple related time series. J-PCMCI+ \citep{gunther2023jpcmci} provides one possible extension for environmental datasets that share partially observed or latent contexts.

\subsection{Methodological, computational, and empirical limitations}

AutoCause adds a decision and evaluation layer around established causal-discovery libraries. It does not replace the algorithms implemented by Tigramite or LiNGAM, and the present evaluation does not support claims of algorithmic superiority over those tools or other causal-discovery frameworks. The contribution lies in recording diagnostic results, applying configurable decision rules, comparing method-specific outputs, and assigning qualified support tiers. For this reason, the original output of each wrapped method remains available rather than being reduced to a single framework-level score.

The identifying assumptions of the wrapped methods remain a primary limitation. The current implementation relies on forms of stationarity, adequate temporal resolution, and a lag window that contains the relevant delays without reducing the effective sample size excessively. Interpolation and sample-size screening provide operational responses to irregular sampling and missing observations, but interpolation can smooth short-lived dynamics or introduce values not directly observed. Likewise, causal-sufficiency diagnostics can flag possible omitted drivers without identifying them. False-positive and false-negative links therefore remain possible even when all recorded diagnostics and safeguards are satisfied.

Edge orientation requires separate interpretation from skeleton recovery. Temporal order determines the direction of lagged links, but some contemporaneous links cannot be uniquely oriented from observational conditional-independence information. Several directed graphs may therefore remain consistent with the same observed dependence structure. Recovering the correct skeleton does not guarantee that every reported direction is correct, and orientation errors may alter the physical interpretation of the inferred process. AutoCause preserves the orientation marks returned by each method, but stronger directional conclusions require domain-specific timing constraints, interventional evidence, or other identifying information.

Computational cost also restricts the current method set. The runtime of CMIknn and LPCMCI increases with the number of variables, record length, lag horizon, and conditioning complexity. The present consensus evaluation excludes interventional methods, feedback-aware approaches, neural causal-discovery models \citep{cheng2024cutsplus,chen2023cuts,cheng2025dycast,yang2026dycausal,zhang2025local}, and methods designed for multiple related datasets. The automatic lag estimator, sample-size thresholds, and recommendation rules also require further sensitivity analysis because the benchmark loops do not independently validate every decision rule.

The empirical evaluation provides only limited evidence for transfer across environmental domains. DGP-Atlas and TimeGraph are synthetic benchmarks, while the CausalRivers analysis uses 30 Bavarian subgraphs of five stations, one calendar year, and one temporal aggregation. The topology-derived reference does not represent rainfall, snowmelt, reservoir regulation, tributary effects, or other processes that may influence discharge. The observed rankings, error rates, and support-tier behavior should therefore not be generalized to other catchments, environmental sensor networks, ecological records, or atmospheric datasets without additional evaluation.

Method-count consensus has a related limitation. On DGP-Atlas and TimeGraph, majority-supported links have the highest observed precision. On CausalRivers, that ordering reverses, showing that several methods can agree on a dependency that is absent from an incomplete reference graph or driven by a shared source of error. The support tiers should therefore be used to prioritize links for further inspection, not as a substitute for a final causal graph.

\section{Conclusions}\label{sec:conclusions}

AutoCause is an open-source decision-support workflow for configuring, running, and documenting causal-discovery methods on multivariate environmental time series. It records method selection, conditional-independence-test selection, lag-window selection, sample-size adequacy, FDR handling, and evidence grading. Preprocessing actions are logged, and defaults can be overridden when physical knowledge or sensitivity analysis supports another setting.

Across DGP-Atlas, TimeGraph, and a topology-derived CausalRivers reference, the evaluated methods show data-regime-dependent graph recovery. Majority-supported links are more precise than single-method links on the synthetic benchmarks, but the ordering breaks on CausalRivers. Consensus support must therefore be read as repeated detection rather than causal validation.

These results support AutoCause as an auditable workflow layer, not a new discovery algorithm or an automatic source of causal truth. Further work should validate decision thresholds independently, isolate component effects, expand environmental testing, and test whether cross-family agreement adds information beyond vote count.

\appendix

\section{Additional analyses of causal evidence and identifiability}
\label{app:causal_evidence}

\setcounter{table}{0}
\renewcommand{\thetable}{\thesection.\arabic{table}}
\renewcommand{\theHtable}{\thesection.\arabic{table}}

\setcounter{figure}{0}
\renewcommand{\thefigure}{\thesection.\arabic{figure}}
\renewcommand{\theHfigure}{\thesection.\arabic{figure}}

\setcounter{equation}{0}
\renewcommand{\theequation}{\thesection.\arabic{equation}}
\renewcommand{\theHequation}{\thesection.\arabic{equation}}

This appendix presents four supplementary analyses that extend the main evaluation without being necessary to interpret its central results. They examine computational cost, the transition from marginal association to conditional causal testing, consistency between empirical performance and method-specific identifiability conditions, and dataset-level surrogate diagnostics. Each analysis follows the metric definitions, graph conventions, and reference-graph qualifications established in the main text.

\subsection{Computational cost}\label{app:runtime}

Table~\ref{tbl:runtime} reports empirical wall-clock time per DGP-Atlas dataset on an Apple M3 Pro with 16~GB RAM. The values combine differences in variable count, data properties, preprocessing, and selected CI test, so they should not be interpreted as a controlled scalability law. The main operational result is that the nonlinear F7 family, which invokes CMIknn for part of the analysis, is substantially slower than the predominantly ParCorr-based families. A controlled scalability analysis that varies $N$, $T$, and $\tau_{\max}$ independently is still required before making general complexity claims.

\begin{table}[pos=t]
\caption{Mean wall-clock time per dataset measured on an Apple M3 Pro with 16\,GB of RAM. Each family contains 10 datasets except F7, which retains 7 after excluding three explosive trajectories. The DGP-Atlas families contain 5--8 variables and series lengths of $T=500$--1000. F7 includes runs in which the conditional-independence test was selected automatically and CMIknn was used for part of the family; the remaining families rely predominantly on ParCorr-based configurations. Values were computed from the per-dataset execution logs included in the released DGP-Atlas results.} \label{tbl:runtime}
\small
\begin{tabular*}{\tblwidth}{@{}lc@{}}
\toprule
Family & Mean time (s) \\
\midrule
F1 (clean VAR, 7 vars)          & 157 \\
F2 (structural breaks, 6 vars)  & 183 \\
F3 (irregular sampling, 5 vars) & 140 \\
F4 (high persistence, 5 vars)   & 161 \\
F5 (latent confounders, 8 vars) & 155 \\
F6 (seasonality, 7 vars)        & 170 \\
F7 (nonlinear, 8 vars)           & 904 \\
F8 (non-Gaussian, 5 vars)       & 161 \\
F9 (mixed violations, 6 vars)   & 125 \\
F10 (extreme cases, 6 vars)     & 231 \\
\bottomrule
\end{tabular*}
\end{table}

\subsection{From marginal association to conditional causal testing}
\label{app:selectivity_progression}

Non-causal reference models are included in the software to show how much structure can be recovered without imposing causal assumptions. In CausalRivers, \citet{stein2025causalrivers} report that simple cross-correlation and reverse-physical baselines can attain scores comparable to those of several causal-discovery methods. \citet{assaad2022survey} likewise use naive approaches as reference points for quantifying the gain associated with stronger causal assumptions, while \citet{runge2018chaos} treats lagged correlation as a minimally restrictive comparator when evaluating PCMCI. The predictive reference extends this comparison to nonlinear structure. Its role follows the distinction drawn by \citet{shmueli2010explain} between explanation and prediction and the argument of \citet{peters2017elements} that predictive accuracy alone does not establish causal direction.

The edge counts expose the difference between marginal association and conditional discovery. As shown in Table~\ref{tbl:results_summary}, on DGP-Atlas, lagged correlation has mean F1\,=\,0.801, recall\,=\,0.988, and precision\,=\,0.684. The high recall follows from testing marginal association across lags without FDR correction across pairs, which admits most true edges but also retains edges that conditional methods would remove. PCMCI+ has mean F1\,=\,0.814 with higher precision and lower recall on the same skeleton endpoint. PCMCI+ has mean F1\,=\,0.814 on the same skeleton endpoint.

On TimeGraph, the lagged-correlation reference attains $\mathrm{F1}=0.608$, recall $=0.722$, and precision $=0.555$, compared with $\mathrm{F1}=0.486$ for PCMCI+ and $0.482$ for VAR-Granger. In several nonlinear and missing-data categories, lagged correlation retains marginal associations that the conditional tests do not detect, while in the clean linear categories (A1, A2) PCMCI+ reaches recall $=1.00$ and precision $=1.00$ and the correlation reference does not. The aggregate advantage of lagged correlation therefore reflects category-level variation rather than a uniform pattern across all data regimes. On CausalRivers, the evaluated methods converge near $\mathrm{F1}\approx0.60$, as high recall is offset by numerous links that are absent from the topology-derived reference. 

These results should not be interpreted as evidence that the non-causal reference models outperform the causal-discovery methods. Instead, they indicate whether substantial associative or predictive structure remains when a causal method reports few links. Lagged correlation and the predictive reference are excluded from the consensus vote because neither provides conditional causal support. 

The patterns we observe are consistent with the assumptions and inferential reach of the methods under study. When their conditional-independence tests are well suited to the data, constraint-based algorithms eliminate indirect links that are still detectable through marginal correlations \citep{runge2018chaos}. Partial correlation performs poorly when the underlying dependence is weakly linear, whereas CMIknn can capture more general functional dependencies, albeit with substantially higher demands on sample size and computation \citep{runge2018conditional}. Linear VAR approaches can still identify the linear component of a structural relationship that also includes nonlinear terms, which aligns with the partial Granger recovery seen in categories that mix linear and polynomial effects \citep{plagborgmoller2021local}. The similar performance scores attained by several methods on the small CausalRivers subgraphs call for a different type of caution, because the topology-based reference is relatively dense given the total number of possible station pairs \citep{petersen2025negative} and does not represent all hydrological mechanisms that generate dependence. Consequently, across the three benchmarks, the difference between non-causal and causal recovery depends on how well method assumptions match the data characteristics, rather than following a uniform performance ranking.

\subsection{Consistency with theoretical identifiability conditions}
\label{app:identifiability}

The three extensions described in Section~\ref{sec:predisc} were developed against DGP-Atlas, so Table~\ref{tbl:audit_recommendations} is an internal consistency check on those extensions rather than an independent validation. It compares the modal causal-audit recommendation on each family with the causal method that attains the highest mean F1 on that family.

\begin{table}[pos=t]
\caption{Causal-audit recommendation versus the highest-mean-F1 causal method on each DGP-Atlas family. Status distinguishes an exact method match from a partial match in which the recommended preprocessing is useful but another causal method has higher mean F1. The random-forest reference is excluded from the target because it is not a causal-discovery method.}\label{tbl:audit_recommendations}
\small
\begin{tabular*}{\tblwidth}{@{}p{2.5cm}p{3.0cm}p{2.5cm}cp{1.6cm}p{1.2cm}@{}}
\toprule
Family & Recommendation & Highest-F1 causal method & F1 & Confidence & Status \\
\midrule
F1 (clean)         & VARLiNGAM         & VARLiNGAM        & 0.95 & 0.85 & Exact \\
F2 (breaks)        & PCMCI+            & PCMCI+            & 0.81 & 0.85 & Exact \\
F3 (irregular)     & PCMCI+ + impute   & VARLiNGAM        & 0.91 & 0.83 & Partial \\
F4 (persistence)   & VARLiNGAM         & VARLiNGAM        & 0.99 & 0.84 & Exact \\
F5 (latent)        & VARLiNGAM         & VAR-Granger      & 0.77 & 0.84 & No \\
F6 (seasonality)   & PCMCI+            & VARLiNGAM        & 0.81 & 0.73 & No \\
F7 (nonlinear)     & PCMCI+            & VAR-Granger      & 0.86 & 0.85 & No \\
F8 (non-Gaussian)  & VARLiNGAM         & VARLiNGAM        & 0.93 & 0.85 & Exact \\
F9 (mixed)         & PCMCI+ + impute   & VAR-Granger      & 0.95 & 0.84 & Partial \\
F10 (extreme)      & VARLiNGAM         & VARLiNGAM        & 0.95 & 0.84 & Exact \\
\bottomrule
\end{tabular*}
\end{table}

A separate comparison asks whether the per-method results follow the identifiability conditions described by \citet{runge2023causal} and \citet{assaad2022survey}. Table~\ref{tbl:theory_vs_obs} maps selected benchmark cells to an assumption-based expectation and the observed best method.

\begin{table}[pos=t]
\caption{Assumption-based expectation and observed mean-F1 pattern on a representative subset of benchmark cells. Cells where the random-forest baseline leads (RF) correspond to regimes in which the evaluated conditional-independence tests have low recall while the predictive reference maintains structure through association alone.}\label{tbl:theory_vs_obs}
\small
\begin{tabular*}{\tblwidth}{@{}p{1.8cm}p{2.2cm}p{5.5cm}p{5.5cm}@{}}
\toprule
Benchmark & Cell & Assumption-based expectation & Observed pattern \\
\midrule
TimeGraph & A1, A2 & PCMCI+ ParCorr (clean linear, sufficient) & PCMCI+ (F1 = 1.00) \\
TimeGraph & B1, B2 & VARLiNGAM/TE (linear test on polynomial fails) & RF (F1 = 0.80) \\
TimeGraph & C1, C2 & VARLiNGAM/TE (seasonality breaks ParCorr) & VARLiNGAM/TE (F1 = 0.80) \\
TimeGraph & D2, D2C & TE (nonparametric on missing data) & RF (sample size too low) \\
DGP-Atlas & F1 (clean) & VARLiNGAM (non-Gaussian preferable to ParCorr) & VARLiNGAM (F1 = 0.95) \\
DGP-Atlas & F3 (irregular) & VARLiNGAM/TE (PCMCI+ assumes regular grid) & VARLiNGAM (F1 = 0.91); PCMCI+ reaches 0.78 after audit-driven imputation \\
DGP-Atlas & F5 (latent) & LPCMCI (admits latent confounders) & VAR-Granger (F1 = 0.77, LPCMCI excluded) \\
DGP-Atlas & F7 (nonlinear) & PCMCI+ CMIknn (theoretical) & VAR-Granger (F1 = 0.86); adaptive PCMCI+ has lower recall \\
DGP-Atlas & F8 (non-Gaussian) & VARLiNGAM (ICA identifiability) & VARLiNGAM (F1 = 0.93) \\
\bottomrule
\end{tabular*}
\end{table}

\subsubsection{DGP-Atlas diagnostic cases}

Table~\ref{tbl:audit_recommendations} shows that the causal-audit recommendation matches the method with highest F1, exactly on five families and partially on two others. The remaining three families (F5, F6, F7) have a \textit{No} status, meaning the recommended method is not the highest scoring one. Each mismatch traces to a different source, ranging from evaluation-target limitations through diagnostic ambiguity to finite-sample behavior of nonlinear tests.

The two partial matches arise from the irregular and mixed families (F3, F9), which exercise both interpolation and CI-test routing. Their released comparisons change more than one setting, so they show the behavior of the audit-driven configuration path rather than an isolated interpolation effect.

Of the three complete mismatches, the latent-variable family (F5) highlights a shortcoming of the evaluation objective rather than of the recommendation mechanism. Its scored graph includes only observed variables, and LPCMCI was not run due to the wall-time constraint. VARLiNGAM is recommended because the risks for nonlinearity, nonstationarity, and irregularity are all low, yet VAR-Granger attains a higher F1 score on this family. As the benchmark does not assess recovery of latent-variable structure, this mismatch stems from the evaluation’s limited scope rather than from an error in the routing logic.

The seasonality family (F6) illustrates a different kind of ambiguity. The module suggests using PCMCI+ with a deseasonalization preprocessing step, yet VARLiNGAM attains the best F1 score (0.813 versus 0.731 for PCMCI+). This difference is relatively small, and the associated recommendation confidence (0.73) is the lowest across all ten families, indicating that the diagnostic metrics do not strongly distinguish these two methods in this setting.

The nonlinear family (F7) highlights a finite-sample constraint. Its structural equations combine linear and polynomial terms, and for the assessed values of $T$, $N$, and $\tau_{\max}$, the RobustParCorr and CMIknn pipelines fail to recover all links. VAR-Granger performs best because it can identify the linear component of the structural equations. This observation is tied to the particular configuration evaluated and should not be interpreted as establishing a general sample-size threshold for CMIknn.

\subsubsection{TimeGraph nonlinear and missing-data cases}

TimeGraph separates nonlinear-test limitations from missing-data limitations. PCMCI+ with ParCorr returns few or no links in polynomial categories where the linear component is weak \citep{runge2018chaos}. Missingness and lag truncation also reduce the effective sample, but the category-level results in Figure~\ref{fig:timegraph_heatmap} show that incomplete series do not fail uniformly.

Transfer entropy tests lags $\{1,2,3,5\}$ per pair and applies Benjamini--Hochberg correction across the four $p$-values. On B1, the ordered values do not satisfy the Benjamini--Hochberg step-up criterion; $\alpha/4=0.0125$ is the first-rank threshold, not a common threshold for every ordered value. The correction therefore retains no tested lag for those weak signals.

Category-level differences are consistent with the inferential targets of the evaluated methods. A linear VAR may recover the linear projection of a relationship that also contains polynomial terms \citep{plagborgmoller2021local}. Linear interpolation may also alter the residual distribution on which VARLiNGAM depends, although the present experiments do not isolate that mechanism \citep{shimizu2006lingam}. The random-forest reference preserves predictive structure in some sparse categories, but it remains excluded from the causal consensus vote.

The tested CMIknn configurations also return low or zero recall on the cubic-quadratic B categories at $T=1000$ and $N=4$. This result is specific to the evaluated signals, conditioning sets, and estimator settings.

\subsubsection{Skeleton and orientation identifiability}\label{sec:direction_limitation}

Edge orientation requires a separate analysis from the skeleton because the endpoints have different identifiability conditions. Under the benchmark's sampling and graph conventions, a positive-lag link is oriented from the variable at $t-\tau$ to the variable at $t$. A contemporaneous link may be identifiable only up to a Markov equivalence class when no collider or orientation rule fixes its direction \citep{spirtes2000causation,pearl2009causality}. PCMCI+ returns an undirected mark (\texttt{o-o}) when several orientations remain compatible with the tested independencies \citep{runge2020discovering}.

AutoCause reports each link at the orientation resolution its method identifies (Section~\ref{sec:methods_taxonomy}). For the constraint-based methods it preserves the Tigramite output convention, in which the array entry indexed by source $i$ at $t-\tau$ and target $j$ at $t$ orients lagged links by time order and retains the contemporaneous marker returned by the algorithm: an arrowhead where an unshielded collider or an orientation rule determines the direction, and an undirected mark (\texttt{o-o}) where it does not. Because the skeleton metric compares unordered pairs, it is insensitive to orientation and serves as the primary comparator (Section~\ref{sec:benchmarks}); orientation is examined separately below, distinguishing links oriented in the generative direction, links left undirected within the equivalence class, and links oriented against it.

The released TimeGraph orientation audit distinguishes correct directions, reversals, and contemporaneous links left undirected. Those outputs are provided under \path{experiments/timegraph_validation/directed_eval/}. They are not combined with skeleton F1 because a fixed-direction score and an equivalence-class-aware score answer different questions.

CausalRivers requires a separate interpretation because its reference specifies upstream-to-downstream links while methods may report lagged, contemporaneous, directed, or undirected outputs. Skeleton recovery remains the primary cross-method endpoint. The directed scoring implementation and outputs are provided in \path{experiments/causalrivers_validation/results_directed/} and \path{experiments/causalrivers_validation/directed_eval.py}.

Table~\ref{tbl:causalrivers_orientation} reports three F1 score levels over the 30 subgraphs: skeleton adjacency, exact directed source-target match, and the identifiable completed partially directed acyclic graph (CPDAG), which credits a correctly oriented link or a contemporaneous link correctly left undirected within the Markov equivalence class. All methods recover the skeleton at comparable F1. Exact-direction F1 is lower for every method. Under the identifiable CPDAG score, PCMCI+ has the highest point estimate because its constraint-based orientation rules leave unorientable contemporaneous links undirected rather than imposing a direction. The random-forest reference is excluded from causal-method rankings.

\begin{table}[pos=t]
\caption{Orientation performance on CausalRivers, reported as mean F1 across 30 subgraphs from three topology classes. Skeleton F1 evaluates unordered adjacency, directed F1 requires an exact source--target match, and identifiable F1 credits either a correctly oriented link or a contemporaneous link that is correctly left undirected within its Markov equivalence class. The random-forest baseline is included only as a non-causal predictive reference and is not ranked with the causal-discovery methods.}
\label{tbl:causalrivers_orientation}
\small
\begin{tabular*}{\tblwidth}{@{}lccc@{}}
\toprule
Method & Skeleton & Directed & Identifiable \\
\midrule
PCMCI+ (ParCorr)   & 0.60 & \textbf{0.40} & \textbf{0.57} \\
VARLiNGAM          & 0.60 & 0.35 & 0.42 \\
Transfer entropy   & 0.60 & 0.29 & 0.34 \\
VAR-Granger        & 0.57 & 0.35 & 0.36 \\
Random forest      & 0.60 & 0.35 & 0.35 \\
\bottomrule
\end{tabular*}
\end{table}

\subsection{Dataset-level surrogate diagnostics}
\label{app:surrogate_falsification}

IAAFT surrogates provide a null model for cross-variable dependence while retaining specified univariate properties \citep{schreiber2000iaaft}. For each variable, the procedure preserves the marginal distribution and approximately preserves the power spectrum. Generating the variables independently disrupts their observed cross-variable dependence. The resulting null does not prove the absence of every causal mechanism.

For each method and dataset, the surrogate edge rate (SER) is the mean fraction of possible directed links returned across 25 surrogate datasets. The separation indicator equals one when the observed edge count exceeds the empirical surrogate edge-count threshold defined in the released falsification script. The reported separation rate is the mean of that indicator across datasets. This dataset-level diagnostic does not test individual links and is not used to assign the consensus-support tiers in Section~\ref{sec:tiering_validation}. The finite-sample rule is recorded in \texttt{experiments/falsification\_validation/results/falsification\_summary.md} and the accompanying script.

The analysis uses a stratified subset of 22 datasets: one DGP from each DGP-Atlas family, six TimeGraph categories, and two CausalRivers subgraphs from each topology class. PCMCI+ uses ParCorr throughout this diagnostic so that the reported surrogate behavior is not mixed with adaptive CI-test selection. Table~\ref{tbl:falsification} reports the resulting SER and separation rates.

\begin{table}[pos=t]
\caption{Dataset-level IAAFT surrogate diagnostics on a 22-dataset stratified subset with 25 surrogates per dataset. SER is the mean fraction of possible directed links under the surrogate null. Separation is the fraction of datasets that pass the finite-sample edge-count rule defined in the released falsification script. Source: \texttt{falsification\_summary.md} and \texttt{falsification\_table.tex}.}\label{tbl:falsification}
\small
\begin{tabular*}{\tblwidth}{@{}lcccccc@{}}
\toprule
& \multicolumn{2}{c}{DGP-Atlas} & \multicolumn{2}{c}{TimeGraph} & \multicolumn{2}{c}{CausalRivers} \\
\cmidrule(lr){2-3} \cmidrule(lr){4-5} \cmidrule(lr){6-7}
Method & SER & Separation & SER & Separation & SER & Separation \\
\midrule
VAR-Granger & \textbf{0.176} & \textbf{90\%} & 0.207 & 33\% & 0.239 & \textbf{100\%} \\
VARLiNGAM & 0.187 & 70\% & 0.177 & 33\% & 0.239 & \textbf{100\%} \\
PCMCI+ (ParCorr) & 0.254 & 80\% & \textbf{0.064} & \textbf{67\%} & \textbf{0.042} & \textbf{100\%} \\
\bottomrule
\end{tabular*}
\end{table}

PCMCI+ has the lowest SER on the evaluated TimeGraph and CausalRivers subsets. The DGP-Atlas subset includes a nonlinear F7 case for which fixed ParCorr is misspecified, but the analysis does not isolate that case's contribution to aggregate SER. A high separation rate means that a method reports more aggregate structure in the observed data than under the IAAFT null. It does not validate individual links or classify links outside the CausalRivers topology reference as physically false.

\section*{Software and data availability}

\subsection*{Software availability}

\begin{description}
\item[Name:] AutoCause
\item[Developer:] Marco Ruiz, ISR-Lisbon, Instituto Superior T\'{e}cnico
\item[Contact:] marco.rueda@tecnico.ulisboa.pt
\item[Year first available:] 2026
\item[Hardware:] No specialized accelerator is required for the small benchmark cases. Runtime and memory increase with the number of variables, record length, lag window, and selected method. Tested on Apple M3 Pro, AMD Ryzen, and ARM-based HPC nodes.
\item[Software:] Python $\geq$ 3.10; tested on Python 3.13. A pinned \texttt{uv.lock} file records the software environment for dependency-level reproducibility.
\item[Program language:] Python
\item[Program size:] Approximately 27,000 Python source lines in the release, measured with \texttt{cloc framework}; \texttt{pytest --collect-only} reports 70 unit and integration tests for discovery methods, sample-size diagnostics, consensus rules, and graph-recovery metrics
\item[Repository:] \url{https://github.com/marcoruizrueda/autocause}. If the repository is private during review, a protected release link is supplied through Editorial Manager.
\item[License:] AGPLv3+
\item[Cost:] Free
\item[Release:] Version 0.2.0, tagged as \texttt{Paper version}. The reviewer package contains the release metadata and software state used for the reported experiments.
\item[Error handling:] Known insufficient-data and assumption-risk conditions are recorded as warnings; method failures are retained in the experiment log and are not converted into detected edges
\item[Documentation:] Usage examples, API reference, and HPC submission scripts are included in the repository
\item[Reproducibility:] The versioned paper release contains experiment scripts, figure-generation scripts, configuration files, and output tables used for the reported analyses. Every numerical value in the paper traces to a specific CSV or script in this release, and a file manifest maps each table and figure to its source.

\end{description}

\subsection*{Data availability statement}

\textbf{Benchmarks.} DGP-Atlas is archived at \url{https://zenodo.org/records/19409395}; TimeGraph is available at \url{https://github.com/hferdous/TimeGraph}; and CausalRivers is available at \url{https://github.com/CausalRivers/causalrivers}. \textbf{Software.} AutoCause is released under AGPLv3+, and causal-audit is available at \url{https://github.com/marcoruizrueda/causal-audit}. If the AutoCause repository is not public during review, the editor and reviewers receive a protected versioned package. \textbf{Reproducibility materials.} That package contains the software environment, experiment configurations, output CSV files, generated tables, figures, and reproduction instructions used for this article.

\section*{Funding}
This work was supported by LARSyS through FCT funding [grant numbers \url{10.54499/LA/P/0083/2020}, \url{10.54499/UIDP/50009/2020}, and \url{10.54499/UIDB/50009/2020}]. The funder had no involvement in study design, data collection, analysis, interpretation, manuscript preparation, or the decision to submit.

\section*{Declaration of competing interests}
The authors declare that they have no known competing financial interests or personal relationships that could have appeared to influence the work reported in this paper.

\printcredits

\section*{Declaration of generative AI and AI-assisted technologies in the manuscript preparation process}
During the preparation of this work, the authors used ChatGPT (OpenAI) and Grammarly to reword and rephrase text, originally written by the authors. After using this tool, the authors reviewed and edited the content as needed and take full responsibility for the content of the publication.

\section*{Acknowledgments}
The authors acknowledge institutional support from ISR-Lisbon, Instituto Superior T\'{e}cnico, and LARSyS.

\bibliographystyle{cas-model2-names}
\bibliography{references}

\end{document}